\documentclass[11pt]{article}

\usepackage[preprint]{acl}

\usepackage{times}
\usepackage{latexsym}

\usepackage[T1]{fontenc}

\usepackage[utf8]{inputenc}

\usepackage{microtype}

\usepackage{inconsolata}

\usepackage{graphicx}
\usepackage{float}

\usepackage{xspace}
\usepackage[table]{xcolor}
\usepackage{tabularx}
\usepackage{array}
\usepackage{multirow}
\usepackage{booktabs}
\usepackage{colortbl}
\usepackage{pifont}

\definecolor{bestcolor}{RGB}{255, 215, 0}      
\definecolor{secondcolor}{RGB}{192, 192, 192}  
\definecolor{sectionbg}{RGB}{240, 240, 245}    
\usepackage{caption}
\usepackage{algorithm}
\usepackage{algpseudocode}
\usepackage{amsmath}
\usepackage[most]{tcolorbox}

\newcommand{\method}{\textsc{GameplayQA}\xspace}

\newcommand{\eg}{\textit{e.g., }}

\usepackage{helvet}
\newcommand{\model}[1]{{{\small\fontfamily{phv}\selectfont{#1}}\xspace}}
\definecolor{gpt_green}{RGB}{22,163,127} 
\definecolor{gemini_blue}{RGB}{81,134,209} 
\definecolor{sonnet35_brown}{RGB}{216, 119, 87} 
\definecolor{qwen_violet}{RGB}{191, 123, 234} 
\definecolor{internvl_blue}{RGB}{0, 153, 255}
\definecolor{seed_teal}{RGB}{0, 180, 180}

\newcommand{\geminiflash}{\model{Gemini-\textcolor{gemini_blue}{3}-Flash}\xspace}
\newcommand{\geminipro}{\model{Gemini-\textcolor{gemini_blue}{3}-Pro}\xspace}

\newcommand{\geminitfpro}{\model{Gemini-\textcolor{gemini_blue}{2.5}-Pro}\xspace}
\newcommand{\geminitfflash}{\model{Gemini-\textcolor{gemini_blue}{2.5}-Flash}\xspace}
\newcommand{\claudehaiku}{\model{Claude-\textcolor{sonnet35_brown}{4.5}-Haiku}\xspace}
\newcommand{\claudesonnet}{\model{Claude-\textcolor{sonnet35_brown}{4.5}-Sonnet}\xspace}

\newcommand{\gptfive}{\model{GPT-\textcolor{gpt_green}{5}}\xspace}

\newcommand{\gptfivemini}{\model{GPT-\textcolor{gpt_green}{5}-mini}\xspace}
\newcommand{\gptfivenano}{\model{GPT-\textcolor{gpt_green}{5}-nano}\xspace}
\newcommand{\qwenEightB}{\model{Qwen-\textcolor{qwen_violet}{3}-VL-8B}\xspace}
\newcommand{\qwenThirtyB}{\model{Qwen-\textcolor{qwen_violet}{3}-VL-30B}\xspace}
\newcommand{\qwenTwoThirtyFiveB}{\model{Qwen-\textcolor{qwen_violet}{3}-VL-235B}\xspace}

\newcommand{\seedone}{\model{Seed-\textcolor{seed_teal}{1.6}}\xspace}
\newcommand{\seedflash}{\model{Seed-\textcolor{seed_teal}{1.6}-Flash}\xspace}
\definecolor{gemma_blue}{RGB}{66, 133, 244}
\newcommand{\gemmaTwentySeven}{\model{Gemma-\textcolor{gemma_blue}{3}-27B}\xspace}
\newcommand{\gemmaTwelve}{\model{Gemma-\textcolor{gemma_blue}{3}-12B}\xspace}
\newcommand{\gemmaFour}{\model{Gemma-\textcolor{gemma_blue}{3}-4B}\xspace}

\title{\method: A Benchmarking Framework for Decision-Dense POV-Synced Multi-Video Understanding of 3D Virtual Agents}

\author{
\textbf{Yunzhe Wang},
\textbf{Runhui Xu},
\textbf{Kexin Zheng},
\textbf{Tianyi Zhang},
\\
\textbf{Jayavibhav Niranjan Kogundi},
\textbf{Soham Hans},
\textbf{Volkan Ustun}
\\[0.5em]
University of Southern California
\\[0.5em]
\small{
\texttt{\{yunzhewa, runhuixu, kexinzhe, tzhang62, jniranja, sohamhan, ustun\}@usc.edu}
}
\\[0.5em]
\small{\textbf{\url{https://hats-ict.github.io/gameplayqa/}}}
}

\begin{document}
\maketitle
\begin{abstract}
  Multimodal LLMs are increasingly deployed as perceptual backbones for autonomous agents in 3D environments, from robotics to virtual worlds.
  These applications require agents to perceive rapid state changes, attribute actions to the correct entities, and reason about concurrent multi-agent behaviors from a first-person perspective, capabilities that existing benchmarks do not adequately evaluate.
  We introduce \method, a framework for evaluating agentic-centric perception and reasoning through  video understanding.
  Specifically, we densely annotate multiplayer 3D gameplay videos at 1.22 labels/second, with time-synced, concurrent captions of states, actions, and events structured around a triadic system of \textbf{Self}, \textbf{Other Agents}, and the \textbf{World}, a natural decomposition for multi-agent environments.
  From these annotations, we refined 2.4K diagnostic QA pairs organized into three levels of cognitive complexity, accompanied by a structured distractor taxonomy that enables fine-grained analysis of where models hallucinate.
  Evaluation of frontier MLLMs reveals a substantial gap from human performance, with common failures in temporal and cross-video grounding, agent-role attribution, and handling the decision density of the game.
  We hope \method stimulates future research at the intersection of embodied AI, agentic perception, and world modeling.
\end{abstract}
\section{Introduction}
\label{sec:introduction}

\begin{figure}[t]
  \centering
  \includegraphics[width=0.99\linewidth]{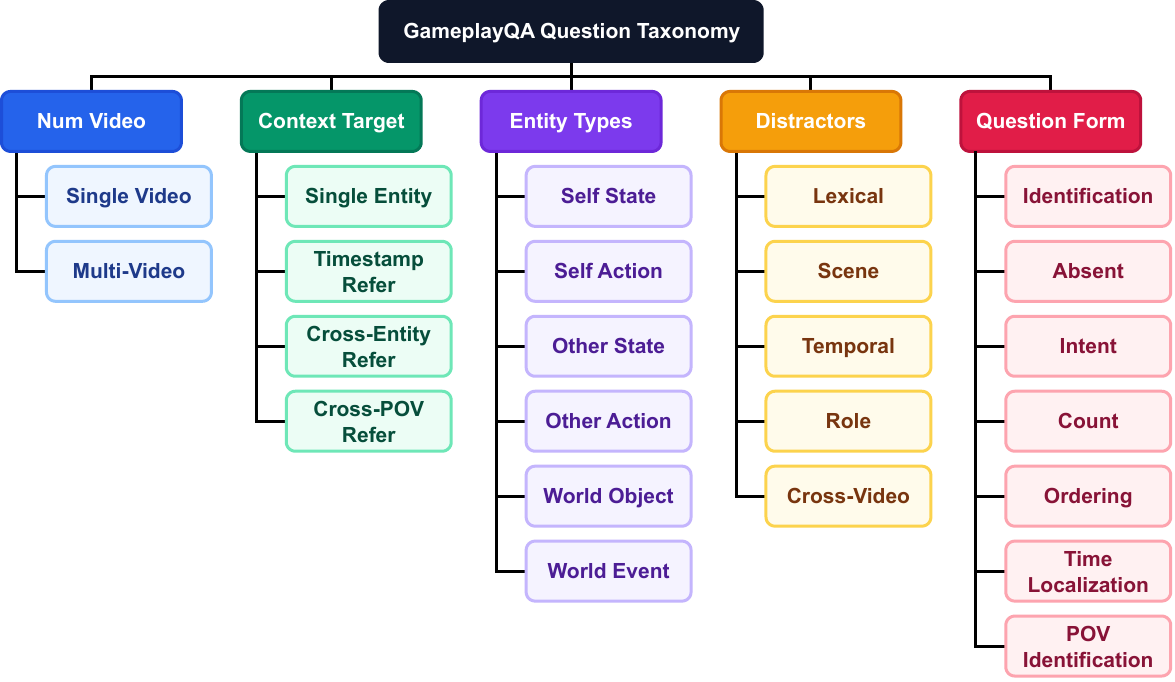}
  \caption{Question taxonomy of \method. Questions are organized along two axes: \textit{Entity} (Self, Other, World) and \textit{Temporal Nature} (Action/State for agents, Object/Event for world), yielding six primitive label types. These primitives compose into 15 task categories across three cognitive levels: single-reference perception (L1), temporal reasoning (L2), and cross-video understanding (L3). See Sec.~\ref{subsec:question_taxonomy} and Table~\ref{tab:question_categories} for details.}
  \label{fig:taxonomy}
\end{figure}

\begin{figure*}[t]
  \centering
  \includegraphics[width=0.99\linewidth]{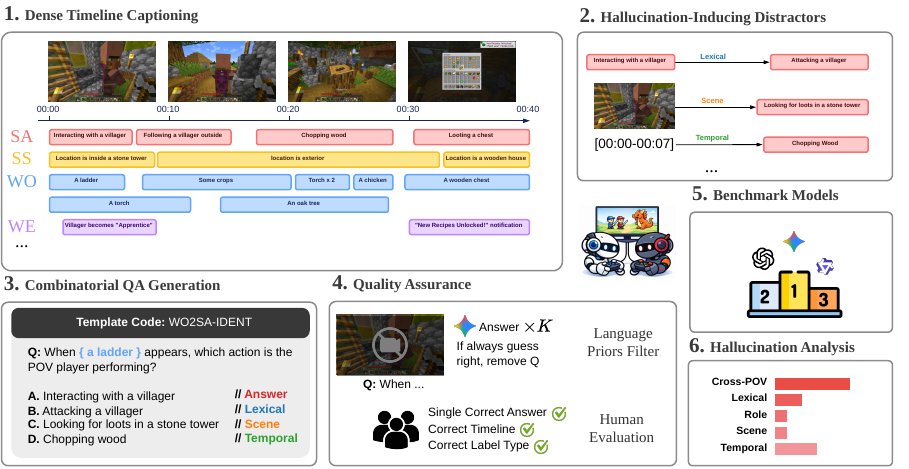}
  \caption{Overview of \method. Gameplay videos undergo (1) dense multi-track temporal captioning on 6 types of target entities (Sec.~\ref{subsec:timeline_captioning}), (2) captioning includes negative labels for hallucination-inducing distractors, and (3) QA pairs are generated through a combinatorial template-based algorithm (Sec.~\ref{subsec:qa_generation}). After (4) quality assurance (Sec.~\ref{subsec:quality_assurance}), the benchmark enables (5) model evaluation with (6) fine-grained hallucination analysis (Sec.~\ref{subsec:hallucination}).}
  \label{fig:framework}
\end{figure*}

\begin{figure*}[t]
  \centering
  \includegraphics[width=0.96\linewidth]{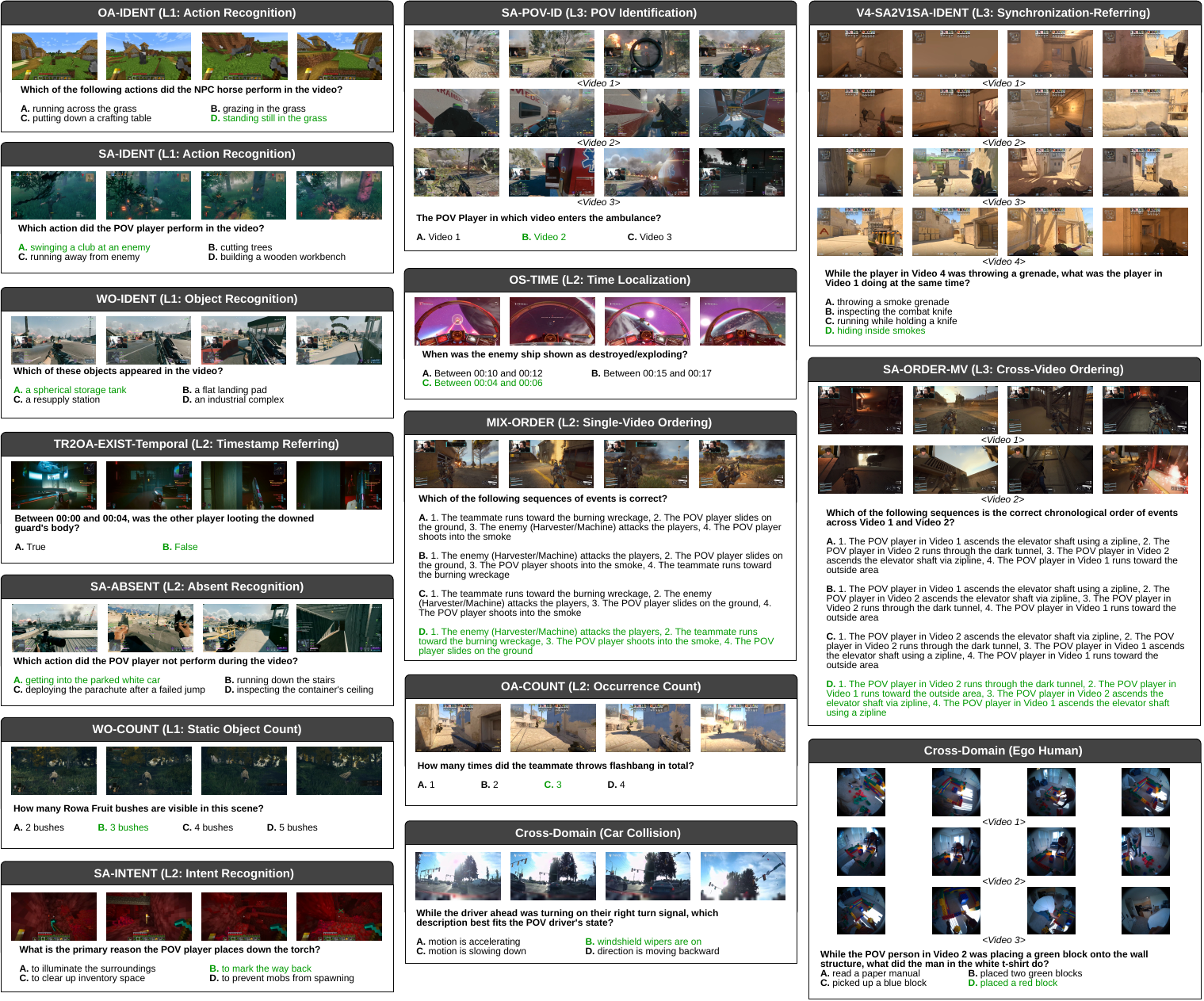}
  \caption{Example questions from \method across different question codes and cognitive levels. Each example shows video frames paired with the corresponding QA pair, illustrating the progression from basic perception (L1) to temporal reasoning (L2) to cross-video understanding (L3). Additional cross-domain examples from car collision and egocentric human activity videos demonstrate the generalizability of the framework.}
  \label{fig:question_example}
\end{figure*}

Recent advances in Multimodal Large Language Models (MLLMs) have demonstrated remarkable capabilities in advanced reasoning, multimodality, and agency \citep{comanici2025gemini,achiam2023gpt,anthropic2024claude45,bai2025qwen3vltechnicalreport}, positioning them as promising decision-making backbones for autonomous agents in Robotics \citep{zitkovich2023rt, geminiroboticsteam2025geminirobotics15pushing}, Computer Use \citep{he2024webvoyager,zhang2025ufo} and 3D virtual agents \citep{simateam2024scalinginstructableagentssimulated,simateam2025sima2generalistembodied,yue2026scaling}. These applications require perception capabilities beyond passive scene description.
Drawing on perspectives from embodied cognition and multi-agent reasoning \citep{hernandez2019survey}, we identify three core requirements for agentic perception in goal-directed environments:
(1) \textbf{dense state-action tracking}: capturing rapid transitions in the agent's own states and actions;
(2) \textbf{other-agent modeling}: reasoning about the behaviors and intentions of other autonomous entities; and
(3) \textbf{environment grounding}: tracking persistent and transient elements of the shared world.

However, current video understanding benchmarks are ill-equipped to diagnose these agentic requirements for three primary reasons. First, the majority of existing evaluation sets suffer from a lack of embodiment and agency grounding \citep{majumdar2024openeqa, yang2025embodiedbench, dang2025ecbench}; they are often composed of slow-paced, passive observations that lack the high-frequency state transitions and dense decision-making loops required to stress-test a model's understanding of intentional action. Second, these benchmarks are largely not hallucination-diagnosable, providing global performance metrics while lacking the granular, multi-faceted annotation needed to identify whether a failure stems from temporal misinterpretation, object fabrication, or role confusion \citep{bai2024hallucination, seth2025egoillusion, tu2025ode}. 
Finally, current protocols exhibit a significant lack of multi-video understanding \citep{peng2025mvuevalmultivideounderstandingevaluation}, focusing almost exclusively on single-viewpoint perception. Multi-video understanding is important in domains such as \textit{sports analytics} leveraging various camera angles and \textit{autonomous driving} requiring information fusion from multiple surround cameras. 
In \textit{esports} and gaming, cross-POV synchronization and collective reasoning, skills that are fundamental to interpreting multi-agent collaboration in interactive 3D spaces \citep{long2024teamcraft, savva2026solaris}, are also crucial.

To bridge this gap, we introduce \method, a comprehensive benchmarking framework, not merely a static evaluation set, but an end-to-end pipeline encompassing structured annotation protocols, automated question generation, and diagnosable error analysis, designed to evaluate the cognitive foundations of agency in 3D virtual environments. We utilize 3D gameplay as a high-density ``cognitive sandbox'' where states and consequences are deterministic and decision-making is fast-paced. 
We meticulously annotate synchronized gameplay videos from 9 multiplayer commercial games at a decision density of $\rho \approx 1.22$ labels/second (Eq.~\ref{eq:decision_density}), using a timeline-based dense captioning mechanism structured around a \textbf{Self}--\textbf{Other}--\textbf{World} entity decomposition. This tripartite schema combined with the properties of 3D gameplay directly addresses the three core agentic requirements identified above: \textit{Self} captures the POV agent's own states and actions for dense state-action tracking; \textit{Other} models external agents' behaviors and intentions; and \textit{World} grounds perception in persistent environmental elements and transient events (Fig.~\ref{fig:sow}). 

Leveraging these annotations, we propose a combinatorial template-based algorithm that generates 2.4K QA pairs organized into a multi-faceted taxonomy spanning three cognitive levels: (1) basic perception, (2) temporal reasoning, and (3) cross-video understanding. The algorithm initially produces 400K candidate pairs and we downsample to 4K to enforce balanced category coverage before quality assurance yields the final set. A key innovation is our structured distractor taxonomy: by categorizing incorrect options as lexical, temporal, or role-based confusions, we can systematically diagnose model hallucination through multiple-choice questions. Evaluation of state-of-the-art MLLMs reveals a  performance gap against human, with models struggling when: (1) the game is fast-paced and decision-dense, (2) questions concern other agents or entities rather than the egocentric player, and (3) cross-video understanding and temporal grounding over long horizons are required. 

In summary, our contributions are threefold:
\begin{itemize}
    \item We introduce an end-to-end benchmarking framework with structured taxonomy, annotation schema, combinatorial QA generation, and diagnosable error analysis, enabling reproducible evaluation pipelines that can scale to new games and domains.
    \item We release a benchmark of 2.4K QA pairs from 9 multiplayer games with synchronized multi-POV videos, filling a critical gap in evaluating the dense, multi-agent perception required for embodied AI.
    \item Benchmarking frontier MLLMs against human evaluation reveals a performance gap, with fine-grained diagnostic analysis through structured distractors revealing that models struggle with fast-paced decision-dense scenarios, other-agent modeling, cross-video synchronization grounding, and temporal reasoning over long horizons.
\end{itemize}

\begin{table*}[t]
  \centering
  \scriptsize
  \begin{tabular}{lcccccccc}
  \toprule
  \textbf{Benchmark} & \textbf{Domain} & \textbf{Agent-Centric} & \textbf{Multi-POV} & \textbf{Diagnostic} & \textbf{Annotation} & \textbf{\#Q} & \textbf{\#Vid} \\
  \midrule
  MVBench~\cite{li2024mvbenchcomprehensivemultimodalvideo} & General & \ding{55} & \ding{55} & \ding{51} & Auto & 4,000 & 4,000 \\
  LongVideoBench~\cite{wu2024longvideobench} & General & \ding{55} & \ding{55} & \ding{51} & Human & 6,678 & 3,763 \\
  MVU-Eval~\cite{peng2025mvuevalmultivideounderstandingevaluation} & General & \ding{55} & \ding{51} & \ding{55} & Human, Auto & 1,824 & 4,959 \\
  MovieQA~\cite{tapaswi2016movieqa} & Movie & \ding{55} & \ding{55} & \ding{55} & Human & 15k & 408 \\
  TVQA~\cite{lei2018tvqa} & TV Shows & \ding{55} & \ding{55} & \ding{55} & Human & 152k & 21.8k \\
  MarioQA~\cite{mun2017marioqa} & Gameplay & \ding{51} & \ding{55} & \ding{55} & Auto & 188k & 13 hrs \\
  PhysGame~\cite{cao2024physgame} & Game Glitches & \ding{55} & \ding{55} & \ding{51} & Human & 880 & 880 \\
  VideoGameQA-Bench~\cite{Taesiri_NeurIPS2025} & Game Glitches & \ding{55} & \ding{55} & \ding{51} & Human, Auto & 4,786 & 1.2k \\
  Ego4D~\cite{grauman2022ego4d} & Daily/Ego & \ding{51} & \ding{55} & \ding{51} & Human & 74k & 3.67k hrs \\
  EgoSchema~\cite{mangalam2023egoschema} & Daily/Ego & \ding{51} & \ding{55} & \ding{51} & Human & 5.1k & 5.1k \\
  EgoIllusion~\cite{seth2025egoillusion} & Daily/Ego & \ding{51} & \ding{55} & \ding{51} & Human, Auto & 8k & 1.4k \\
  \midrule
  \textbf{GameplayQA (Ours)} & \textbf{Gameplay} & \ding{51} & \ding{51} & \ding{51} & Human, Auto & 2,365 & 100 \\
  \bottomrule
  \end{tabular}
  \caption{Comparison of relevant video understanding benchmarks.}
  \label{tab:benchmark_comparison}
\end{table*}
  
\begin{table*}[t]
\centering
\scriptsize
  
\begin{tabular}{llllcc}
\toprule
\textbf{Context Scope} & \textbf{Task} & \textbf{Description} & \textbf{Example Question Codes} & \textbf{\#Q} & \textbf{Dur.} \\
\midrule
\multirow{5}{*}{\shortstack[l]{\textbf{Single}\\\textbf{Reference}\\ (L1, 469)}} 
& Action Recognition & Identify or verify existence of self \& others' actions & \texttt{SA-IDENT}, \texttt{OA-EXIST}, ... & 162 & 10.0s \\
& State Recognition & Identify or verify existence of self \& others' states & \texttt{SS-IDENT}, \texttt{OS-EXIST}, ... & 147 & 10.1s \\
& Object Recognition & Identify or verify existence of world objects in scene & \texttt{WO-IDENT}, \texttt{WO-EXIST} & 70 & 9.3s \\
& Event Recognition & Identify world events occurring in the environment & \texttt{WE-IDENT} & 61 & 8.4s \\
& Static Object Count & Count static objects present in the scene & \texttt{WO-COUNT} & 29 & 21.3s \\
\midrule
\multirow{7}{*}{\shortstack[l]{\textbf{Temporal}\\(L2, 1383)}} 
& Cross-Entity Referring & Link one entity to another (X2Y reasoning) & \texttt{SA2SS-IDENT}, \texttt{WO2SS-EXIST}, ... & 423 & 23.0s \\
& Timestamp Referring & Given time range [t1-t2], identify what entity exists & \texttt{TR2SS-IDENT}, \texttt{TR2SA-IDENT}, ... & 81 & 24.3s \\
& Time Localization & Locate exact timestamp when an event occurred & \texttt{SA-TIME}, \texttt{WE-TIME}, ... & 281 & 28.4s \\
& Absence Recognition & Identify actions/states that did not occur over a timespan & \texttt{SA-ABSENT}, \texttt{SS-ABSENT}, ... & 195 & 38.9s \\
& Occurrence Count & Count how many times an action/event happened & \texttt{SA-COUNT}, \texttt{OA-COUNT}, \texttt{WE-COUNT} & 75 & 26.4s \\
& Ordering & Determine temporal order sequence of actions/events & \texttt{SA-ORDER}, \texttt{OA-ORDER}, \texttt{MIX-ORDER} & 180 & 32.6s \\
& Intent Identification & Identify underlying intent or goal behind actions & \texttt{SA-INTENT}, \texttt{OA-INTENT} & 148 & 23.0s \\
\midrule
\multirow{3}{*}{\shortstack[l]{\textbf{Cross-Video}\\(L3, 513)}} 
& Sync-Referring & Link corresponding entities across synchronized videos & \texttt{V1-SA2-V2OA}, \texttt{V1-WO2-V2SS}, ... & 207 & 94.7s \\
& Cross-Video Ordering & Determine event order sequence across multiple videos & \texttt{SA-ORDER-MV}, \texttt{MIX-ORDER-MV}, ... & 117 & 110.0s \\
& POV Identification & Identify who performed what action in which video & \texttt{SA-POV-ID}, \texttt{OA-POV-ID}, ... & 189 & 91.6s \\
\bottomrule
\end{tabular}
\caption{Definition of 15 categories of cross-entity referring reasoning questions in \method, including the number of questions and average video duration.}
  \label{tab:question_categories}
  \end{table*}

\section{Related Work}
\label{sec:related_work}

\paragraph{Multimodal Large Language Models}
Recent progress in MLLMs has significantly expanded the ability of AI systems to perceive and reason over visual inputs~\citep{comanici2025gemini,achiam2023gpt,anthropic2024claude45,bai2023qwen}. Many recent MLLMs have been proposed to be video-native for video understanding~\citep{cheng2024videollama,comanici2025gemini,li2024llama}. These systems can process extended visual streams; however, they remain prone to hallucination, including fabricating objects, misinterpreting temporal dynamics, and confusing causal relationships~\citep{bai2024hallucination,he2025evaluating,tu2025ode,seth2025egoillusion}. 

\paragraph{Video Understanding Benchmarks}
Video understanding benchmarks have evolved from early action recognition datasets toward evaluations emphasizing temporal reasoning, spatial grounding, and long-context comprehension. General video QA benchmarks such as MVBench~\citep{li2024mvbenchcomprehensivemultimodalvideo}, LongVideoBench~\citep{wu2024longvideobench}, Video-MME~\citep{fu2025video}, and MVU-Eval~\citep{peng2025mvuevalmultivideounderstandingevaluation} assess multimodal models on fine-grained temporal perception and multi-step inference. Domain-specific benchmarks target narrative understanding in movies and TV shows~\citep{tapaswi2016movieqa,lei2018tvqa}. Egocentric benchmarks including Ego4D~\citep{grauman2022ego4d}, EgoSchema~\citep{mangalam2023egoschema}, ECBench~\citep{dang2025ecbench}, and EgoIllusion~\citep{seth2025egoillusion} evaluate first-person video understanding and hallucination detection. Embodied QA benchmarks such as OpenEQA~\citep{majumdar2024openeqa} and EmbodiedBench~\citep{yang2025embodiedbench} ground reasoning in physical environments. In the video game domain, MarioQA~\citep{mun2017marioqa} pioneered event-centric QA on 2D platformer videos, while recent works explored the feasibility of using MLLMs to detect video game graphics glitches, including GlitchBench~\citep{taesiri2024glitchbench}, VideoGameQA-Bench~\citep{Taesiri_NeurIPS2025}, and PhysGame~\citep{cao2024physgame}.

\section{The GameplayQA Framework}
\label{sec:framework}

We collected 3D multiplayer gameplay footage from 9 commercial games spanning diverse genres (see Appendix~\ref{sec:dataset_stats} for the full game list). Videos were sourced from YouTube, Twitch streams, and existing datasets~\citep{wang2025x}. For games requiring synchronized multi-POV footage, we identified groups of streamers who played together in the same match and downloaded their individual recordings, then manually aligned them to construct temporally synchronized multi-video sets.

This section details how we obtain the benchmark from these raw videos: defining a question taxonomy (Section~\ref{subsec:question_taxonomy}), annotating via timeline captioning on synchronized multi-POV videos (Section~\ref{subsec:timeline_captioning}), generating QA pairs through a combinatorial template-based algorithm (Section~\ref{subsec:qa_generation}), and applying quality assurance procedures (Section~\ref{subsec:quality_assurance}). The final benchmark contains 2.4K QA pairs, generated from 2709 caption true labels and 1586 distractor labels.

\subsection{Question Taxonomy}
\label{subsec:question_taxonomy}

Our question taxonomy (Figure~\ref{fig:taxonomy}) is built upon a \textbf{six-primitive label system} that categorizes observable events along two axes: \textit{Agent} (Self, Other, World) and \textit{Temporal Nature} (Action/State for agents, Object/Event for world).

\paragraph{Entity Types.}
We organize perception in interactive 3D environments around three entity categories (Figure~\ref{fig:sow}): \textbf{Self} (the POV agent), \textbf{Other} (external entities such as teammates, enemies, or NPCs), and \textbf{World} (the shared environment). This Self--Other--World decomposition naturally aligns with multi-agent reinforcement learning frameworks and agent-based modeling paradigms  \citep{sutton1998reinforcement,busoniu2008comprehensive}, where agents must simultaneously track their own state, model other agents' behaviors, and respond to environmental dynamics (Illustration in Fig.~\ref{fig:sow}). 
For each entity category, we distinguish between dynamic and static properties: \textbf{Self-Action} (SA) captures what the player \textit{does} (shooting, jumping, reloading), while \textbf{Self-State} (SS) captures the player's \textit{condition} (health, ammo, equipped weapon). Similarly, \textbf{Other-Action} (OA) and \textbf{Other-State} (OS) track other agents. 
The \textit{World} category is divided into \textbf{World-Object} (WO), referring to static or interactive items such as supply crates and vehicles, and \textbf{World-Event} (WE), which includes dynamic events like explosions or game notifications. 
This labeling system enables hallucination analysis of model error rates by entity type (see Sec.~\ref{subsec:hallucination}).

\paragraph{Task Categories.}
We organize questions into 15 task categories across three cognitive levels; question examples, category sizes, and average video durations are summarized in Table~\ref{tab:question_categories}. 
\textbf{Level 1 (Single Reference)} tests basic perception: recognizing actions, states, objects, and events within a single video segment. These tasks include action recognition (\eg \textit{``What did the player do?''}), state recognition (\eg \textit{``What was the player's health?''}), object recognition, event recognition, and static object counting. \textbf{Level 2 (Temporal)} introduces temporal reasoning that requires grounding answers to specific time windows. Tasks include cross-entity referring (\eg \textit{``When the player jumped, what was their health?''}), timestamp referring, time localization, absence recognition (identifying what did \textit{not} occur), occurrence counting, temporal ordering, and intent identification. \textbf{Level 3 (Cross-Video)} extends reasoning across synchronized multi-POV footage, testing sync-referring (\eg \textit{``When POV1 was reloading, what did POV2 do?''}), cross-video ordering, and POV identification. This hierarchy progressively tests from basic perception to complex multi-perspective temporal reasoning. Figure \ref{fig:question_example} provides typical example questions covering the task categories.

\paragraph{Distractor Taxonomy.}
A key contribution of \method is its structured distractor taxonomy, which enables fine-grained diagnosis of why models hallucinate. We categorize incorrect options by their relationship to the ground truth. \textbf{Lexical} distractors are text-based variants of the correct option, generated by changing the subject, using antonyms, or altering object attributes. \textbf{Scene} distractors are vision-based options listing plausible events that did not actually occur in the video. \textbf{Temporal} distractors refer to events that \textit{did} happen, but outside the queried time window. \textbf{Role} distractors swap the agent attribution (\eg attributing other agents' actions to the POV player). \textbf{Cross-Video} distractors refer to events from other synchronized videos, applicable only to multi-video questions. 
By analyzing the error rates for each distractor type, we can pinpoint failure modes in temporal grounding, agent attribution, or semantic understanding.

\subsection{Multi-Video Timeline Captioning}
\label{subsec:timeline_captioning}

We employ \textbf{dense multi-track timeline captioning} where each of the six entity types (SA, SS, OA, OS, WO, WE) is treated as an independent annotation track (See Figure \ref{fig:interface_single} and Figure \ref{fig:interface_multi} for screenshots of labeling interface). Labels within and across tracks can overlap temporally, enabling concurrent event capture (\eg a player action (SA) occurring while their health state (SS) changes during a world event (WE)). Figure \ref{fig:framework} visualizes this process, where the object label ``a ladder'' is temporally referred to ask a question regarding the player's action at the same time. For multi-POV videos, we synchronize timelines across perspectives, enabling cross-video temporal alignment.

\paragraph{Decision Density.}
We operationalize decision density as the temporal frequency of semantic labels such as actions, states, and events that constitute the necessary information stream for an agent's planning and reaction loop. Formally, we define the density metric $\rho$ as:
\begin{equation}
  \rho = \frac{N_{\text{labels}}}{T_{\text{seconds}}}
  \label{eq:decision_density}
\end{equation}
Across our benchmark, 2,709 true labels span a total of 2,219.41 seconds of annotated footage, yielding $\rho \approx 1.22$ labels/second. Table~\ref{tab:label_density} (Appendix~\ref{sec:dataset_stats}) shows the per-type breakdown, reflecting the predominance of self-centric observations in first-person gameplay. 
This high-frequency labeling regime sets \method apart from passive video benchmarks and underscores the inherent difficulty of temporal grounding tasks in our experiments.

The annotation process follows a two-stage human-in-the-loop workflow. In the first stage, \geminipro generates candidate labels (3,632 predictions) and distractors (1,678 predictions). Four graduate student annotators then verify and refine these candidates: 31.1\% of predicted labels were deleted, 42.7\% were edited (with 61.9\% requiring caption changes and 42.2\% requiring temporal boundary adjustments), and 26.2\% were accepted without modification. Additionally, 7.6\% of the final label set were added entirely by annotators to capture events missed by the model. In the second stage, a separate annotator reviews all labels, making further adjustments to approximately 12\% of labels. Detailed annotation protocol and annotator statistics are provided in Appendix~\ref{sec:annotation_protocol}.

\subsection{Combinatorial QA Generation}
\label{subsec:qa_generation}

We generate questions through a \textbf{combinatorial template-based algorithm} that instantiates question templates by systematically combining verified labels across five orthogonal dimensions: number of videos, context target, entity type, distractor type, and question form, as summarized in Table~\ref{tab:question_categories} and Table~\ref{tab:question_taxonomy}. For each combination, the algorithm selects a ground-truth label as the correct answer and populates the remaining options with distractors drawn from the corresponding distractor pool, enabling fine-grained diagnosis of model failure modes. Complete templates are listed in Appendix~\ref{sec:question_templates}. Optionally, an LLM paraphrasing step is applied to reword the templated questions into more natural phrasing without altering their meaning or answer.

The algorithm initially produces 399,214 candidate QA pairs. Sync-Referring, Cross-Entity Referring, Timestamp Referring, and Ordering types dominate due to their combinatorial nature, so we strategically downsample to 4K questions to enforce balanced category coverage and avoid long-tail bias. After quality assurance described in Section~\ref{subsec:quality_assurance}, this yields the final 2,365 gold-standard pairs.

\begin{table*}[ht]
  \centering
  \scriptsize
  \setlength{\tabcolsep}{3pt}
  \renewcommand{\arraystretch}{1.1}
  \begin{tabular}{lc ccccc ccccccc ccc}
  \toprule
  \multirow{2}{*}{Model} & \multirow{2}{*}{All}
  & \multicolumn{5}{c}{L1 (Single Ref.)}
  & \multicolumn{7}{c}{L2 (Temporal)}
  & \multicolumn{3}{c}{L3 (Cross-Video)} \\
  \cmidrule(lr){3-7}
  \cmidrule(lr){8-14}
  \cmidrule(lr){15-17}
  & & ActRec & StaRec & ObjRec & EvtRec & SOC
  & X-Ent & TsRef & TimLoc & AbsRec & OccCnt & Order & Intent
  & SyncRef & X-VOrd & POV-ID \\
  \midrule
  Human & 80.5 & 80.0 & 80.0 & 100.0 & 75.0 & 100.0 & 84.2 & 100.0 & 76.9 & 83.3 & 62.5 & 77.8 & 57.1 & 88.9 & 77.8 & 100.0 \\
  \rowcolor{sectionbg}
  \multicolumn{17}{c}{\textit{Proprietary Models}} \\
  GPT-5 & 67.0 & \cellcolor{bestcolor}79.0 & \cellcolor{bestcolor}70.7 & 70.0 & 68.9 & 48.3 & \cellcolor{secondcolor}71.6 & 70.4 & 45.9 & \cellcolor{bestcolor}86.2 & \cellcolor{secondcolor}62.7 & 78.3 & 54.1 & 72.0 & \cellcolor{secondcolor}60.7 & 54.0 \\
  GPT-5 Mini & 62.7 & 70.4 & 67.3 & 68.6 & 72.1 & 34.5 & 67.6 & 66.7 & 47.0 & 79.0 & 33.3 & 72.8 & 50.0 & 72.0 & 43.6 & 58.7 \\
  GPT-5 Nano & 49.3 & 61.7 & 60.5 & 70.0 & 72.1 & 37.9 & 57.7 & 65.4 & 33.5 & 64.6 & 17.3 & 35.0 & 41.9 & 49.8 & 29.1 & 42.9 \\
  Gemini 2.5 Pro & \cellcolor{bestcolor}71.3 & 69.1 & \cellcolor{secondcolor}68.0 & 70.0 & \cellcolor{secondcolor}80.3 & 34.5 & \cellcolor{bestcolor}74.5 & \cellcolor{secondcolor}77.8 & \cellcolor{bestcolor}65.1 & 82.1 & 38.7 & \cellcolor{bestcolor}82.8 & 59.5 & \cellcolor{bestcolor}81.0 & \cellcolor{bestcolor}65.0 & \cellcolor{bestcolor}85.7 \\
  Gemini 3 Flash & \cellcolor{secondcolor}68.2 & \cellcolor{secondcolor}71.6 & 65.3 & \cellcolor{secondcolor}75.7 & 68.9 & 24.1 & 70.7 & \cellcolor{bestcolor}80.2 & \cellcolor{secondcolor}64.4 & \cellcolor{secondcolor}83.6 & 32.0 & \cellcolor{secondcolor}78.9 & 62.8 & \cellcolor{secondcolor}76.3 & 52.1 & \cellcolor{secondcolor}60.3 \\
  Gemini 2.5 Flash & 63.7 & 69.8 & 59.2 & 71.4 & 72.1 & 31.0 & 65.0 & 69.1 & 60.5 & 76.9 & 34.7 & 72.2 & 60.8 & 72.9 & 50.4 & 51.3 \\
  Claude 4.5 Sonnet & 51.3 & 62.3 & 49.7 & 70.0 & 65.6 & 48.3 & 57.9 & 50.6 & 34.9 & 68.2 & 41.3 & 42.2 & 61.5 & 47.8 & 30.8 & 46.0 \\
  Claude 4.5 Haiku & 41.8 & 46.9 & 52.4 & 60.0 & 60.7 & 51.7 & 41.8 & 53.1 & 26.0 & 53.3 & 24.0 & 36.1 & 46.6 & 41.1 & 29.9 & 38.6 \\
  Seed 1.6 & 61.8 & 75.9 & 63.3 & \cellcolor{bestcolor}77.1 & 73.8 & 51.7 & 70.4 & 65.4 & 44.1 & 78.5 & 42.7 & 69.4 & 60.1 & 57.0 & 41.9 & 47.6 \\
  Seed 1.6 Flash & 56.5 & 66.9 & 56.1 & 72.1 & 74.6 & 50.0 & 65.5 & 67.9 & 30.9 & 68.8 & 38.4 & 41.5 & 63.1 & 61.7 & 48.2 & 44.2 \\
  \rowcolor{sectionbg}
  \multicolumn{17}{c}{\textit{Open-Source Models}} \\
  Qwen3 VL 235B & 63.8 & 71.0 & 59.9 & 70.0 & \cellcolor{secondcolor}80.3 & 55.2 & 68.6 & 76.5 & 54.4 & 80.0 & 50.7 & 72.8 & \cellcolor{secondcolor}63.5 & 66.7 & 31.6 & 49.2 \\
  Qwen3 VL 30B & 60.8 & 68.5 & 60.5 & 74.3 & \cellcolor{bestcolor}82.0 & \cellcolor{secondcolor}58.6 & 65.2 & \cellcolor{secondcolor}77.8 & 47.7 & 79.5 & \cellcolor{bestcolor}65.3 & 66.7 & 56.8 & 55.1 & 30.8 & 47.1 \\
  Qwen3 VL 8B & 57.8 & 68.5 & 56.5 & 74.3 & 72.1 & \cellcolor{bestcolor}62.1 & 63.6 & 75.3 & 46.3 & 73.3 & 52.0 & 57.2 & \cellcolor{bestcolor}64.9 & 48.3 & 27.4 & 45.5 \\
  Gemma 3 27B & 48.0 & 55.6 & 54.4 & 58.6 & 60.7 & 44.8 & 57.4 & 64.2 & 29.2 & 66.2 & 32.0 & 28.3 & 50.7 & 46.4 & 29.9 & 46.0 \\
  Gemma 3 12B & 43.7 & 53.1 & 48.3 & 65.7 & 59.0 & 31.0 & 52.5 & 54.3 & 26.7 & 54.9 & 9.3 & 27.2 & 50.0 & 50.2 & 24.8 & 39.7 \\
  Gemma 3 4B & 42.9 & 46.9 & 42.9 & 64.3 & 63.9 & 24.1 & 49.6 & 54.3 & 26.0 & 57.4 & 9.3 & 27.2 & 46.6 & 58.5 & 23.9 & 37.6 \\
  \midrule
  \textit{All Models} & 56.9 & 64.8 & 58.4 & 69.5 & 70.4 & 43.0 & 62.5 & 66.8 & 42.7 & 72.0 & 36.5 & 55.5 & 55.8 & 59.8 & 38.8 & 49.6 \\
  \bottomrule
  \end{tabular}
  \caption{Model performance across task categories. \colorbox{bestcolor}{Gold} = best, \colorbox{secondcolor}{Silver} = second best. L1: ActRec (Action Recognition), StaRec (State Recognition), ObjRec (Object Recognition), EvtRec (Event Recognition), SOC (Static Object Count). L2: X-Ent (Cross-Entity Referring), TsRef (Timestamp Referring), TimLoc (Time Localization), AbsRec (Absence Recognition), OccCnt (Occurrence Count), Order (Ordering), Intent (Intent Identification). L3: SyncRef (Sync-Referring), X-VOrd (Cross-Video Ordering), POV-ID (POV Identification).}
  \label{tab:main_results}
\end{table*}

\subsection{Quality Assurance}
\label{subsec:quality_assurance}

\paragraph{Language Prior Filtering.} Template-based generation can introduce language priors that allow models to guess answers without visual grounding. To mitigate this, we apply a \textbf{blind filtering} procedure: for each generated question, we query \geminiflash with $k=3$ trials using only the question text (no video). Questions where the model consistently achieves high accuracy are flagged as potentially biased and removed from the benchmark. This ensures that remaining questions require genuine video understanding rather than exploiting statistical regularities in question phrasing.

\paragraph{Human Evaluation.} To validate generation quality, we evenly sampled 120 questions covering all question types for human evaluation. Annotators assessed two criteria: (1) the video contains exactly one correct answer among the options, and (2) the question adheres to the semantics defined by its question code (\eg an IDENT question truly requires identification). For questions where annotators disagreed, we held discussion meetings to reach consensus; when no agreement could be reached, we resolved through majority voting. During this process, 8\% of questions were flagged as faulty due to issues such as excessive similarity between multiple options or misaligned temporal boundaries, which is consistent with the annotation error propagation discussed in our limitations (Section~\ref{sec:limitations}).

\section{Experiments}
\label{sec:experiments}

We evaluate both open-source and proprietary MLLMs. \textbf{Open-source:} Qwen3-VL Series\citep{bai2025qwen3vltechnicalreport}, Gemma 3 Series \citep{gemmateam2025gemma3technicalreport}. \textbf{Proprietary:} GPT-5 Series \citep{openai2025gpt5}, Claude 4.5 \citep{anthropic2024claude45} (Sonnet, Haiku), Gemini Series \citep{comanici2025gemini}, and Seed 1.6 \citep{guo2025seed1}.

\paragraph{Evaluation Setup.} We evaluate all models in a zero-shot setting using accuracy as the metric. For video-native models (Gemini, Seed), we input the entire video directly. For frame-based models, we sample frames at 1 FPS up to 32 frames; for videos longer than 32 seconds, we uniformly sample 32 frames across the duration. Videos are resized such that the longer side is 720p while preserving aspect ratio. Although models are instructed to output a single letter, they sometimes produce full sentences or explanations; we use \gptfivemini as an LLM judge to extract the selected option. Detailed inference settings are in Appendix~\ref{sec:appendix:inference_settings_and_providers}; evaluation prompt templates in Appendix~\ref{sec:appendix:prompt_templates}.

\subsection{Main Results}
\label{subsec:main_results}

Table~\ref{tab:main_results} summarizes model performance across all task categories. Among all models evaluated, Gemini 2.5 Pro attains the highest overall accuracy (71.3\%), followed by Gemini 3 Flash (68.2\%) and GPT-5 (67.0\%), yet a substantial gap to human performance (80.5\%) persists. We highlight two key findings below.

\paragraph{Consistent degradation across cognitive levels.} Averaged across all models, accuracy drops steadily from L1 Single-Reference (61.2\%) to L2 Temporal (56.0\%) to L3 Cross-Video (49.4\%). This trend validates that the three-level hierarchy of \method successfully stratifies task difficulty, with temporal grounding and multi-POV reasoning remaining substantially more challenging than basic visual perception.

\paragraph{Counting and Cross-Video Ordering are the hardest tasks.} Two tasks emerge as clear bottlenecks. Occurrence Count (OccCnt) averages only 36.5\% across models, making it the hardest L2 task. This suggests that tracking event recurrences over time, which demands sustained temporal attention across frames, remains beyond the reach of current models. Cross-Video Ordering (X-VOrd) averages 38.8\%, the lowest among L3 tasks, with several models dropping to around 30\%, indicating severe difficulty in aligning temporal events across perspectives. Together, these results suggest that precise temporal tracking, whether within a single video or across multiple perspectives, remains a fundamental weakness of current video-language architectures.

\begin{table}[ht]
  \centering
  \scriptsize
  \setlength{\tabcolsep}{4pt}
  \renewcommand{\arraystretch}{1.1}
  \begin{tabular}{lc cccccc}
  \toprule
  Model & All & SA & SS & OA & OS & WO & WE \\
  \midrule
  \rowcolor{sectionbg}
  \multicolumn{8}{c}{\textit{Proprietary Models}} \\
  Gemini 2.5 Pro & \cellcolor{bestcolor}71.3 & \cellcolor{bestcolor}75.2 & \cellcolor{secondcolor}73.3 & \cellcolor{secondcolor}65.6 & \cellcolor{bestcolor}72.0 & \cellcolor{secondcolor}71.7 & \cellcolor{bestcolor}74.0 \\
  Gemini 3 Flash & \cellcolor{secondcolor}68.2 & \cellcolor{secondcolor}70.5 & 73.0 & \cellcolor{bestcolor}65.6 & \cellcolor{secondcolor}64.8 & 70.5 & 68.4 \\
  GPT-5 & 67.0 & 66.0 & \cellcolor{bestcolor}73.5 & 60.8 & 67.0 & \cellcolor{bestcolor}71.6 & \cellcolor{secondcolor}69.1 \\
  Gemini 2.5 Flash & 63.7 & 65.1 & 66.6 & 60.8 & 60.1 & 67.6 & 64.9 \\
  GPT-5 Mini & 62.7 & 64.4 & 68.6 & 57.6 & 60.4 & 66.8 & 63.8 \\
  Seed 1.6 & 61.8 & 60.3 & 65.8 & 57.6 & 62.3 & 67.0 & 65.6 \\
  Seed 1.6 Flash & 56.5 & 56.9 & 59.3 & 56.7 & 49.6 & 63.0 & 63.5 \\
  Claude 4.5 Sonnet & 51.3 & 51.6 & 53.0 & 50.9 & 46.9 & 57.8 & 53.6 \\
  GPT-5 Nano & 49.3 & 47.1 & 59.9 & 47.7 & 50.3 & 55.9 & 51.4 \\
  Claude 4.5 Haiku & 41.8 & 41.0 & 39.3 & 38.1 & 44.0 & 44.1 & 45.7 \\
  \rowcolor{sectionbg}
  \multicolumn{8}{c}{\textit{Open-Source Models}} \\
  Qwen3 VL 235B & 63.8 & 63.6 & 67.4 & 59.2 & 58.8 & 70.8 & 70.2 \\
  Qwen3 VL 30B & 60.8 & 57.6 & 62.7 & 57.1 & 60.1 & 66.2 & 67.8 \\
  Qwen3 VL 8B & 57.8 & 53.7 & 59.9 & 56.8 & 56.3 & 65.9 & 63.4 \\
  Gemma 3 27B & 48.0 & 48.4 & 53.5 & 48.3 & 47.8 & 54.6 & 49.7 \\
  Gemma 3 12B & 43.7 & 45.2 & 51.4 & 42.1 & 44.7 & 47.3 & 46.4 \\
  Gemma 3 4B & 42.9 & 43.4 & 49.6 & 43.2 & 40.3 & 52.2 & 47.7 \\
  \midrule
  \textit{All Models} & 56.8 & 56.5 & 61.0 & 54.0 & 55.4 & 62.0 & 60.2 \\
  \bottomrule
  \end{tabular}
  \caption{Model performance by entity category. \colorbox{bestcolor}{Gold} = best, \colorbox{secondcolor}{Silver} = second best. SA: Self-Action, SS: Self-State, OA: Other-Action, OS: Other-State, WO: World-Object, WE: World-Event.}
  \label{tab:entity_category}
\end{table}

\subsection{Error Source Analysis}
\label{subsec:hallucination}

\begin{figure}[t]
  \centering
  \includegraphics[width=\columnwidth]{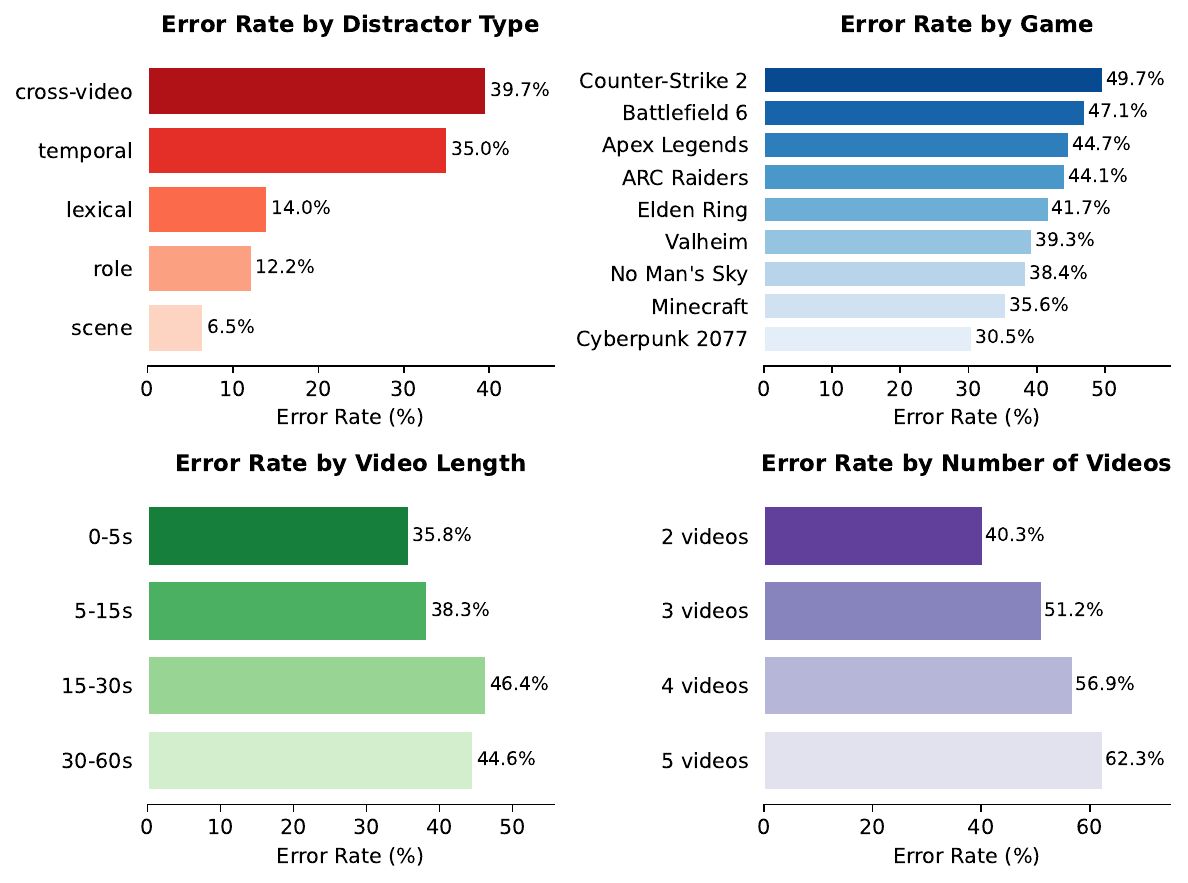}
  \caption{Error rate analysis across four dimensions. \textbf{Top-left}: Cross-video and temporal distractors cause the most errors. \textbf{Top-right}: Fast-paced shooters (CS2, Battlefield) are hardest. \textbf{Bottom-left}: Error increases with video length. \textbf{Bottom-right}: Error scales with number of synchronized videos.}
  \label{fig:error_analysis}
\end{figure}

We conduct a fine-grained error analysis to identify systematic failure modes by entity category. Table~\ref{tab:entity_category} reveals that \textbf{World-Object (WO)} recognition is the easiest category (62.0\% aggregate), while recognizing \textbf{Other Agents} proves substantially harder, for example Other-Action (OA) at 54.0\% and Other-State (OS) at 55.4\% represent an 8-point gap compared to world objects. This suggests MLLMs struggle with other agent attribution in multi-agent scenes.

We further plot error rates along four criteria: distractor type in EXIST questions (True/False), game name, video length, and number of synchronized videos. Figure~\ref{fig:error_analysis} reveals three key insights. First, models are primarily confused by cross-video and temporal distractors, while scene distractors are the easiest, indicating that models handle static visual input better than temporal and cross-video reasoning. Second, game pace strongly predicts difficulty: competitive shooters with rapid state transitions (Counter-Strike, Battlefield, Apex Legends) rank as the top three hardest games compared to slower exploration titles, validating that \textit{decision-dense environments} pose fundamentally harder challenges. Third, both temporal extent and multi-POV complexity compound errors, as longer clips and additional synchronized perspectives each generally degrade performance monotonically.

\subsection{Language Prior and Temporal Ablation}
\label{subsec:ablation}

To disentangle the contributions of visual grounding and temporal reasoning, we conduct ablation studies on \gptfivemini under three degraded input conditions: (1) \textbf{No Video}, where only the question text is provided; (2) \textbf{Random Frame}, where a single randomly selected frame replaces the full video; and (3) \textbf{Shuffled Frames}, where the original frames are presented in random order. Results are shown in Table~\ref{tab:ablation}.

\begin{table}[ht]
  \centering
  \small
  \begin{tabular}{lcccc}
  \toprule
  \textbf{Condition} & \textbf{All} & \textbf{L1} & \textbf{L2} & \textbf{L3} \\
  \midrule
  Baseline (Full) & 62.7 & 67.2 & 61.9 & 60.6 \\
  No Video & 29.4 & 36.0 & 29.1 & 24.2 \\
  Random Frame & 41.7 & 52.9 & 40.9 & 33.7 \\
  Shuffled Frames & 54.8 & 63.1 & 52.6 & 53.4 \\
  \bottomrule
  \end{tabular}
  \caption{Ablation study on \gptfivemini with degraded visual inputs. Performance drops indicate the contribution of video content and temporal ordering.}
  \label{tab:ablation}
\end{table}

\begin{table*}[ht]
  \centering
  \scriptsize
  \setlength{\tabcolsep}{3pt}
  \renewcommand{\arraystretch}{1.1}
  \begin{tabular}{lc ccccc ccccccc ccc}
  \toprule
  \multirow{2}{*}{Model} & \multirow{2}{*}{All}
  & \multicolumn{4}{c}{L1 (Single Ref.)}
  & \multicolumn{7}{c}{L2 (Temporal)}
  & \multicolumn{3}{c}{L3 (Cross-Video)} \\
  \cmidrule(lr){3-6}
  \cmidrule(lr){7-13}
  \cmidrule(lr){14-16}
  & & ActRec & StaRec & ObjRec & EvtRec
  & X-Ent & TsRef & TimLoc & AbsRec & OccCnt & Order & Intent
  & SyncRef & X-VOrd & POV-ID \\
  \midrule
  Gemini 2.5 Pro   & \cellcolor{bestcolor}66.2 & 61.1 & \cellcolor{secondcolor}86.7 & \cellcolor{secondcolor}83.3 & \cellcolor{bestcolor}100.0 & 62.0 & \cellcolor{secondcolor}80.0 & 52.6 & 72.2 & 40.0 & 50.0 & \cellcolor{bestcolor}66.7 & \cellcolor{bestcolor}76.7 & 25.0 & \cellcolor{bestcolor}55.6 \\
  Gemini 2.5 Flash & \cellcolor{secondcolor}62.0 & \cellcolor{bestcolor}72.2 & 73.3 & 66.7 & \cellcolor{secondcolor}80.0  & \cellcolor{secondcolor}64.0 & 80.0 & 36.8 & 61.1 & \cellcolor{secondcolor}40.0 & \cellcolor{bestcolor}83.3 & 55.6 & \cellcolor{secondcolor}70.0 & \cellcolor{secondcolor}50.0 & 38.9 \\
  GPT-5 Mini       & 61.0 & 55.6 & 66.7 & \cellcolor{bestcolor}100.0 & \cellcolor{secondcolor}80.0 & \cellcolor{bestcolor}66.0 & 70.0 & \cellcolor{secondcolor}63.2 & \cellcolor{bestcolor}77.8 & 20.0 & \cellcolor{secondcolor}66.7 & 66.7 & 53.3 & 50.0 & 27.8 \\
  Qwen3 VL 235B    & 59.2 & 50.0 & \cellcolor{bestcolor}73.3 & \cellcolor{bestcolor}100.0 & \cellcolor{bestcolor}100.0 & 66.0 & 80.0 & 36.8 & \cellcolor{secondcolor}72.2 & 20.0 & 50.0 & \cellcolor{secondcolor}66.7 & 50.0 & 25.0 & \cellcolor{secondcolor}44.4 \\
  \bottomrule
  \end{tabular}
  \caption{Cross-domain generalization results on real-world ego-centric videos (autonomous driving + multi-human collaboration, 213 questions). \colorbox{bestcolor}{Gold} = best, \colorbox{secondcolor}{Silver} = second best. Column headers follow the same conventions as Table~\ref{tab:main_results}.}
  \label{tab:cross_domain}
\end{table*}

The No Video condition drops accuracy the largest amount (33.3\%), confirming that \method requires genuine visual grounding and cannot be solved by language priors alone. The Random Frame condition recovers only partial performance (+12.3\% over No Video), indicating that static visual content provides useful context but cannot substitute for temporal dynamics. Shuffled Frames achieves near-baseline L1 performance (63.1\% vs.\ 67.2\%) but degrades substantially on L2 (52.6\% vs.\ 61.9\%) and L3 (53.4\% vs.\ 60.6\%), showing that temporal ordering is critical for reasoning tasks but less so for basic perception.

\subsection{Cross-Domain Generalization}
\label{subsec:cross_domain}

To validate that our framework generalizes beyond gameplay to broader single-agent and multi-agent ego-centric settings, we conducted a small-scale transfer experiment by applying the identical benchmarking pipeline to two real-world domains: (1) dashcam collision videos from the Nexar dataset~\citep{moura2025nexar}, and (2) synchronized ego-centric videos of humans collaboratively assembling Lego from the Ego-Humans benchmark~\citep{khirodkar2023ego}. The only pipeline adjustment required was renaming the default actor from ``player'' to domain-appropriate labels such as ``person'' or ``driver''.

Across 4 videos ($\sim$113 seconds total), our automated pipeline generated 5,463 initial questions; following the same downsampling and quality-assurance protocol as the main benchmark, we produced a test set of \textbf{213 questions} at a label density of $\rho = 0.50$ labels/second, lower than gameplay ($\rho = 1.22$), reflecting the slower decision pace of real-world activities.

Table~\ref{tab:cross_domain} shows that performance trends mirror those of the main benchmark: Gemini 2.5 Pro leads overall (66.2\%), models degrade progressively from L1 to L3, and Occurrence Count and Cross-Video Ordering remain the hardest tasks. The lower label density confirms that real-world videos progress at a slower decision pace than gameplay, yet the relative difficulty ordering across models and task categories is preserved. These results demonstrate that our benchmarking framework generalizes to real-world spatiotemporal tasks with only minimal domain-specific adjustments.

\section{Conclusion}
\label{sec:conclusion}

We presented \method, an end-to-end benchmarking framework that uses densely annotated, synchronized multi-POV gameplay videos to evaluate agentic perception in decision-dense 3D environments. Built on a Self--Other--World entity decomposition and a three-level cognitive hierarchy, the framework refines 2.4K diagnostic QA pairs whose structured distractors pinpoint where models hallucinate. Evaluation of 16 frontier MLLMs reveals steady performance degradation from basic perception to temporal reasoning to cross-video understanding, with models particularly failing on other-agent attribution, temporal grounding, and fast-paced decision-dense scenarios. Cross-domain experiments on autonomous driving and ego-centric human collaboration confirm that the pipeline generalizes with minimal adaptation, preserving difficulty and model rankings across domains. We hope \method drives progress toward MLLMs capable of reliable perception and reasoning in dynamic, multi-agent worlds.

\section*{Limitations}
\label{sec:limitations}

While \method includes intent identification as a proxy for understanding goal-directed behavior, our framework does not cover decision reasoning questions such as \textit{``What is the best action to take at this moment?''}. Answering such questions would require estimating expected rewards or action values from raw video observations, which is a capability that remains an open research challenge, as it demands not only perception but also learning implicit reward structures and world dynamics from uncurated, in-the-wild footage. Additionally, intent identification is inherently more subjective than recognizing physical actions or states, which occasionally results in multiple defensible answers; during human evaluation, approximately 8\% of questions were flagged as having ambiguous ground-truth labels (Section~\ref{subsec:quality_assurance}). Nevertheless, because anticipating intent is a critical capability for planning agents, we believe this category provides vital diagnostic signal and should be preserved.

\paragraph{Annotation Cost and Error Propagation.}
The dense labeling process underlying \method is extremely labor-intensive and susceptible to human error. Annotators must track over 100 labels and distractors per video, repeatedly watching the same footage while switching cognitive focus across different entity types and temporal windows. On average, labeling a 30-second video clip requires 25--35 minutes. More critically, the combinatorial QA generation algorithm reuses labels across multiple questions, meaning that a single labeling error, whether in timestamp boundaries, entity type, or description content, can propagate to multiple erroneous questions. However, because the QA generation algorithm is deterministic and template-based, perfectly accurate annotations would yield perfectly correct questions; i.e. the source of benchmark noise is annotation error, not algorithmic error. While our quality assurance procedures mitigate this risk, some annotation noise inevitably remains in the benchmark.

\section*{Acknowledgments}

The authors acknowledge the use of Large Language Models for assistance with proofreading and grammar checking. All content was reviewed, edited, and approved by the human authors, who take full responsibility for the final manuscript. The project or effort depicted was or is sponsored by the U.S. Army Combat Capabilities Development Command -- Soldier Centers under contract number W912CG-24-D-0001. The content of the information does not necessarily reflect the position or the policy of the Government, and no official endorsement should be inferred.

\bibliography{ref}

\clearpage
\appendix
\definecolor{plcaption}{RGB}{0, 102, 204}      
\definecolor{plother}{RGB}{255, 140, 0}        

\newcommand{\plcaption}[1]{\texttt{\textcolor{plcaption}{\{#1\}}}}
\newcommand{\plother}[1]{\texttt{\textcolor{plother}{\{#1\}}}}

\section{GameplayQA Question Taxonomy}
\label{app:question_taxonomy}

\begin{figure}[h]
    \centering
    \includegraphics[width=0.7\linewidth]{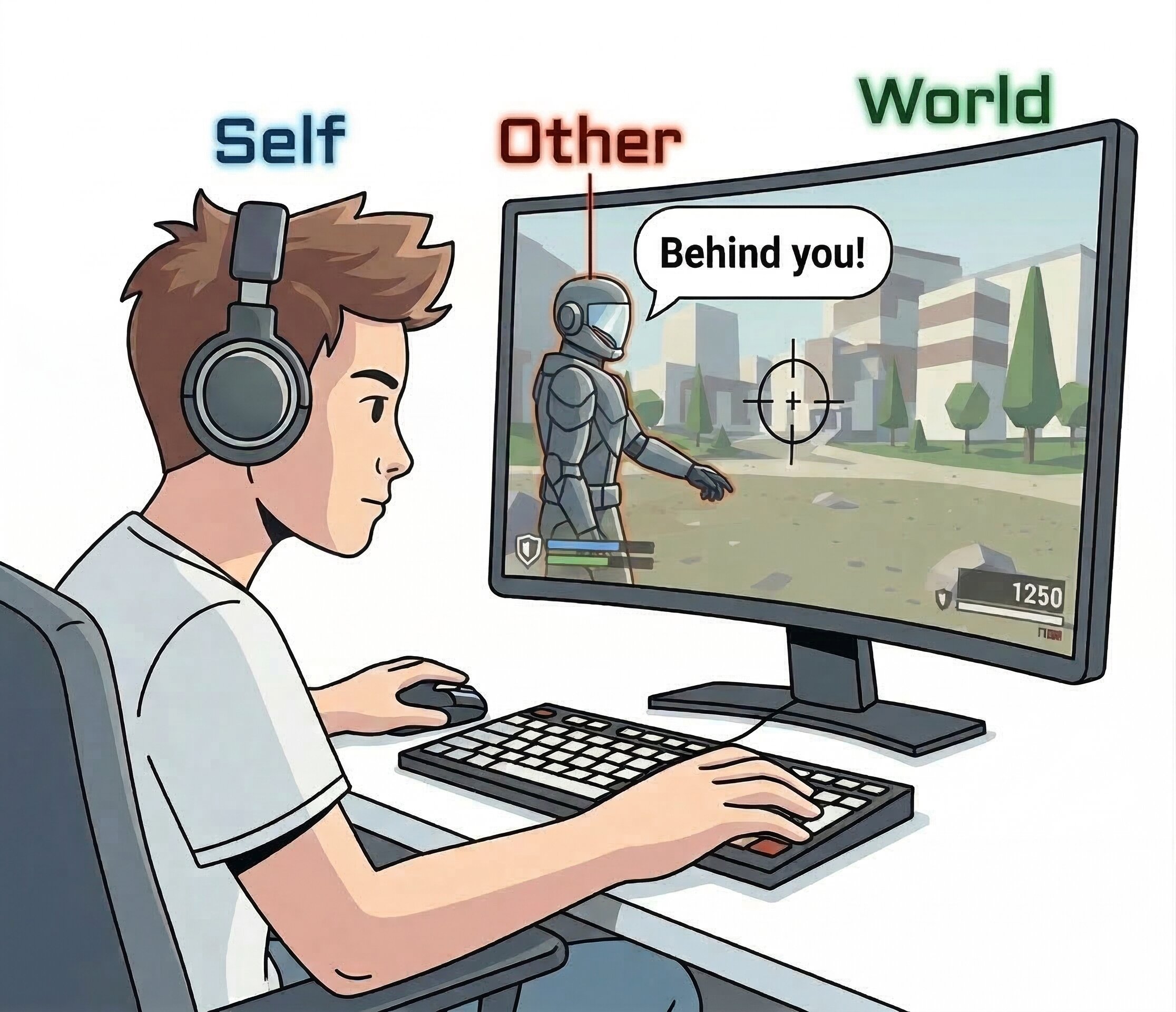}
    \caption{Illustration of the \textbf{Self--Other--World} framework. A first-person player (Self) perceives a teammate (Other) issuing a warning, set against the surrounding game environment (World). These three perspectives define the entity types used in our question taxonomy.}
    \label{fig:sow}
\end{figure}

Table~\ref{tab:question_taxonomy} summarizes the GameplayQA question taxonomy. Each question is defined by five orthogonal dimensions. For each category, we provide a short description and representative example questions. Figure~\ref{fig:sow} illustrates the Self--Other--World concept.

\begin{table*}[t]
\centering
\small
\setlength{\tabcolsep}{8pt}
\renewcommand{\arraystretch}{1.4}
\begin{tabular}{p{2.2cm}p{2.1cm}p{5.2cm}p{5.8cm}}
\toprule
\textbf{Dimension} & \textbf{Category} & \textbf{Description} & \textbf{Example Question} \\
\midrule

\multirow{2}{*}{\textit{Number of Videos}}
& Single Video
& Inputs a single video
& What action did the player perform? \\

& Multi-Video
& Inputs multiple videos
& When POV1 player was reloading, which action did POV2 player perform? \\

\cmidrule{1-4}

\multirow{4}{*}{\textit{Context Target}}
& Summative
& (Single-Video Only) Requires aggregation over a temporal segment.
& Which of the following best summarizes the player's actions during the video? \\

& Timestamp \newline Referring
& Refers to a specific moment or interval
& At [02:45 - 02:52], what action did the player perform? \\

& Target Entity \newline Referring
& Refers to the temporal moment of the 6 entity types.
& When the player was reloading, what action did the teammate perform? \\

& Cross-Video \newline Referring
& (Multi-Video Only) Explicitly references a video index.
& Which POV player reload their weapon? \\

\cmidrule{1-4}

\multirow{6}{*}{\textit{Entity Type}}
& Self-Action
& Action performed by the POV player \eg shooting, reloading, using item, etc
& Which action did the POV player performed? \\

& Self-State
& State or status of the POV player \eg health, inventory, equipped weapon, etc
& What was the player's health status at that time? \\

& Other-Action
& Action performed by another player \eg teammate, enemy, NPC, etc
& What action did the enemy perform during the fight? \\

& Other-State
& State of another player
& Was the teammate downed during the encounter? \\

& World-Object
& Environment objects or landmarks \eg tree, supply crate, building, cars, etc
& How many supply crates are visible in the scene? \\

& World-Event
& Environmental or system-level events or notifications \eg explosion, enemy downed, achievement notification, etc
& Did an explosion occur during this segment? \\

\cmidrule{1-4}

\multirow{5}{*}{\textit{Distractor Type}}
& Lexical
& Textually similar but incorrect descriptions
& Did the player reload instead of switching weapons? \\

& Scene
& Plausible but nonexistent events
& Did a vehicle explode in this area? \\

& Temporal
& Real events outside the question context
& Did the explosion occur before the firefight? \\

& Role
& Correct event but wrong agent
& Did the enemy reload their weapon? \\

& Intent
& Alternative plausible motivations
& Did the player reload to prepare for a long fight? \\

\cmidrule{1-4}

\multirow{5}{*}{\textit{Question Form}}
& Identification
& Select the correct answer from options
& Which of the following actions occurred? \\

& Existence
& Binary true or false question
& Did an explosion occur in this clip? \\

& Absent
& Identify what did not occur
& Which event did not happen during healing? \\

& Intent
& Ask why an action was performed
& Why did the player reload their weapon? \\

& Count
& Ask for quantities or frequencies
& How many enemies appeared in the scene? \\

\bottomrule
\end{tabular}
\caption{GameplayQA question taxonomy. Each question is defined by a combination of dimensions, enabling systematic coverage of perception, temporal reasoning, and cross-video reasoning.}
\label{tab:question_taxonomy}
\end{table*}

\section{Model Details and Inference Settings}
\label{sec:model_details}
\label{sec:appendix:inference_settings_and_providers}

This section provides details about the inference settings and providers used to run the benchmark. Table~\ref{tab:inference_config} summarizes the inference configurations for all tested models.

\begin{table*}
\centering
\begin{tabular}{llll}
\toprule
\textbf{Model Name} & \textbf{Version} & \textbf{Inference Provider} & \textbf{Platform} \\
\midrule
\gptfive & gpt-5 & OpenAI & \href{https://openai.com/api/}{OpenAI} \\
\gptfivemini & gpt-5-mini & OpenAI & \href{https://openai.com/api/}{OpenAI} \\
\gptfivenano & gpt-5-nano & OpenAI & \href{https://openai.com/api/}{OpenAI} \\
\geminitfpro & gemini-2.5-pro & Google & \href{https://aistudio.google.com/}{Google AI Studio} \\
\geminiflash & gemini-3-flash & Google & \href{https://aistudio.google.com/}{Google AI Studio} \\
\geminitfflash & gemini-2.5-flash & Google & \href{https://aistudio.google.com/}{Google AI Studio} \\
\claudesonnet & claude-4.5-sonnet & Amazon Bedrock & \href{https://openrouter.ai/}{OpenRouter} \\
\claudehaiku & claude-4.5-haiku & Amazon Bedrock & \href{https://openrouter.ai/}{OpenRouter} \\
\seedone & bytedance-seed/seed-1.6 & Seed & \href{https://openrouter.ai/}{OpenRouter} \\
\seedflash & bytedance-seed/seed-1.6-flash & Seed & \href{https://openrouter.ai/}{OpenRouter} \\
\qwenTwoThirtyFiveB & qwen/qwen3-vl-235b-a22b-instruct & Fireworks &  \href{https://openrouter.ai/}{OpenRouter}\\
\qwenThirtyB & qwen/qwen3-vl-30b-a3b-instruct & Fireworks &  \href{https://openrouter.ai/}{OpenRouter}\\
\qwenEightB & qwen/qwen3-vl-8b-instruct & Alibaba &  \href{https://openrouter.ai/}{OpenRouter}\\
\gemmaTwentySeven & google/gemma-3-27b-it & Chutes & \href{https://openrouter.ai/}{OpenRouter} \\
\gemmaTwelve & google/gemma-3-12b-it & Chutes & \href{https://openrouter.ai/}{OpenRouter} \\
\gemmaFour & google/gemma-3-4b-it & Chutes & \href{https://openrouter.ai/}{OpenRouter} \\
\bottomrule
\end{tabular}
\caption{Inference configurations for tested models. All models are evaluated using the listed inference providers.}
\label{tab:inference_config}

\vspace{1em}

\begin{tabular}{@{}ll@{}}
\toprule
\textbf{Model Name} & \textbf{Default Reasoning Mode} \\
\midrule
\gptfive{} & Balanced (medium) \\
\gptfivemini{} & Balanced (medium) \\
\gptfivenano{} & Balanced (medium) \\
\geminitfpro{} & High (dynamic) \\
\geminiflash{} & High (dynamic) \\
\geminitfflash{} & High (dynamic) \\
\claudesonnet{} & Standard (extended thinking off) \\
\claudehaiku{} & Standard (extended thinking off) \\
\bottomrule
\end{tabular}
\caption{Default reasoning effort for tested models. All models are evaluated using their default reasoning settings.}
\label{tab:reasoning_effort}
\end{table*}

\paragraph{Video Input Strategy.}
For \textbf{video-native models} (\geminitfpro, \geminiflash, \geminitfflash), we input the entire video directly without frame sampling. For \textbf{non-video-native models} (GPT, Claude, Qwen, Gemma), we sample frames at 1 FPS up to a default of 32 frames; for videos longer than 32 seconds, we uniformly sample 32 frames across the duration. All videos are resized such that the longer side is 720p while preserving aspect ratio.

\paragraph{Special Cases.}
\textbf{Qwen 30B and 235B:} Due to provider (Fireworks) limitations, we cap the total number of frames at 30 for these models. For multi-video questions, the 30 frames are evenly split across all input videos (\eg 10 frames per video for 3 synchronized videos).

\textbf{Seed 1.6 and Seed 1.6 Flash:} Although these models support video-native input, we experienced high instability when calling the API with raw video. We therefore resort to 32-frame sampling for these models to ensure consistent evaluation.

\textbf{Gemini 3-Pro:} We did not benchmark \geminipro due to strict rate limits (250 API calls per day) at the time this paper was written, which made full benchmark evaluation infeasible.

\paragraph{Reasoning Effort Settings.}
For models that support configurable reasoning modes, we use the default reasoning effort settings provided by each API. Table~\ref{tab:reasoning_effort} summarizes the default reasoning modes for each model family.

\section{Dataset Statistics}
\label{sec:dataset_stats}

\subsection{Games Selection and Data Source}
\label{sec:games}

GameplayQA includes footage from 9 commercially released games spanning diverse genres:

\begin{itemize}
    \item \textbf{Single-POV Games:} Minecraft, Apex Legends, No Man's Sky, Elden Ring, Cyberpunk 2077, Valheim
    \item \textbf{Multi-POV Synchronized Games:} Counter-Strike 2~\citep{wang2025x}, Battlefield 6, Arc Raiders
\end{itemize}

For multi-POV games, synchronized footage was obtained either from existing datasets or by identifying groups of Twitch streamers who played together in the same match and manually aligning their individual recordings.

\subsection{Label Distribution}

Table~\ref{tab:label_density} reports the per-type breakdown of the 2,709 true labels annotated across 2,219.41 seconds of footage, corresponding to a decision density of $\rho \approx 1.22$ labels/second.

\begin{table}[H]
  \centering
  \small
  \setlength{\tabcolsep}{5pt}
  \begin{tabular}{lcc}
  \toprule
  \textbf{Label Type} & \textbf{Count} & \textbf{Share} \\
  \midrule
  Self-Action (SA)   & 658 & 24.3\% \\
  Self-State (SS)    & 729 & 26.9\% \\
  Other-Action (OA)  & 160 &  5.9\% \\
  Other-State (OS)   & 190 &  7.0\% \\
  World-Event (WE)   & 417 & 15.4\% \\
  World-Object (WO)  & 555 & 20.5\% \\
  \midrule
  \textbf{Total}     & \textbf{2,709} & \textbf{100\%} \\
  \bottomrule
  \end{tabular}
  \caption{Per-type label distribution ($\rho \approx 1.22$ labels/s over 2,219.41 s of annotated footage).}
  \label{tab:label_density}
\end{table}

\begin{figure*}[p]
  \centering
  \includegraphics[width=\textwidth,height=\textheight,keepaspectratio]{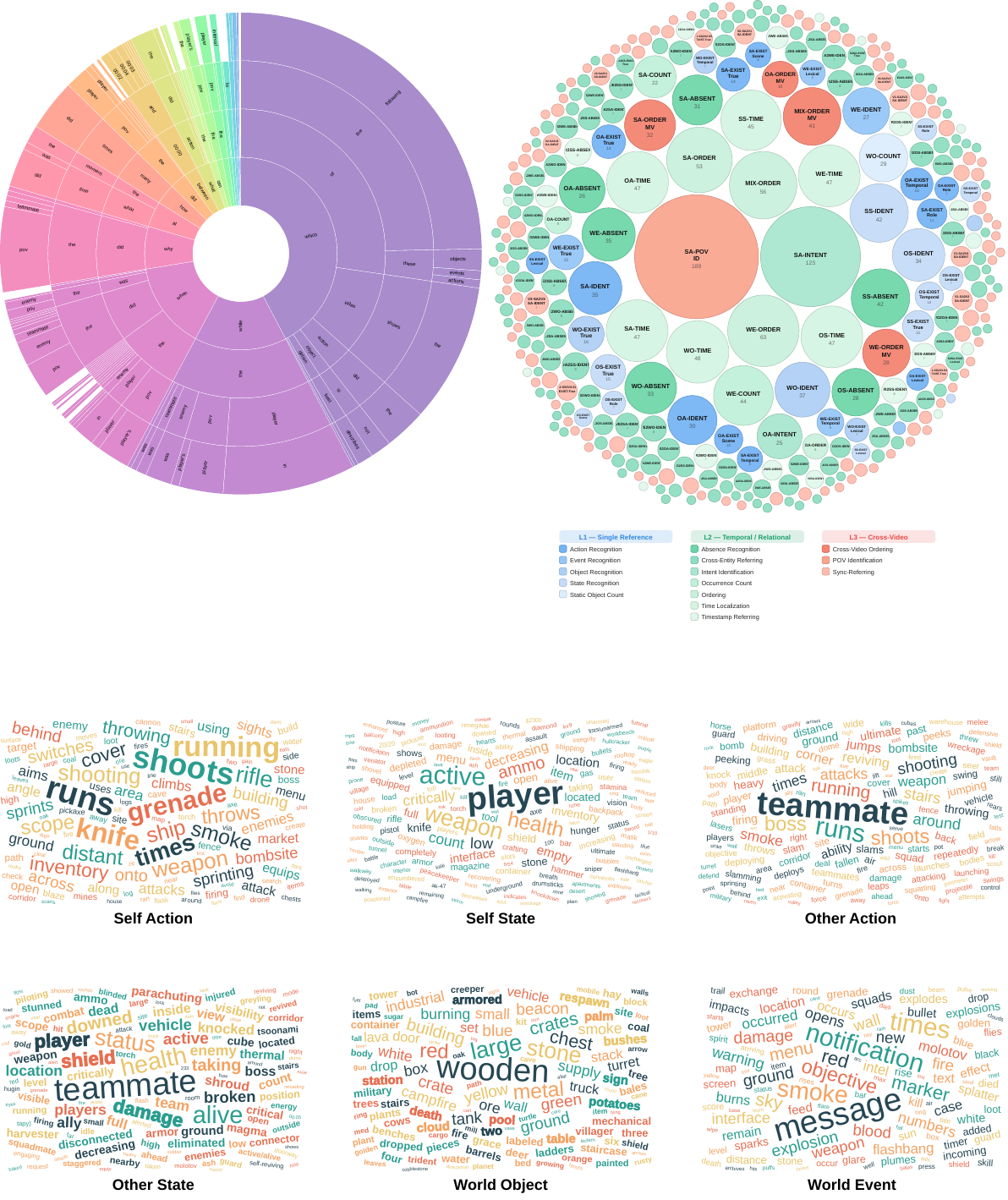}
  \caption{
  \textbf{Top left:} Distribution of the first four words of the questions. The arc length indicates the frequency.
  \textbf{Top right:} Distribution of question codes, categorized by count and task type (using color for cognitive level).
  \textbf{Bottom:} Word cloud visualizing question terms by entity type.
  }
  \label{fig:dataset_statistics}
\end{figure*}

Figure~\ref{fig:dataset_statistics} provides a broader view of the dataset regarding the distribution of question opening phrases, the question codes distribution by count and task type, and word distribution by entity type.

\section{Prompt Templates}
\label{sec:appendix:prompt_templates}

\begin{tcolorbox}[title = {Single-Video Question Answering}]
Watch the video carefully and answer the following multiple choice question:

\texttt{<frame\_1> <frame\_2> ... <frame\_32>}

Q: \texttt{<question>}

\texttt{<options>}

Please select the correct answer from the options. Answer with the letter directly.

Your answer:
\end{tcolorbox}

\begin{tcolorbox}[title = {Multi-Video Question Answering}]
Watch the video carefully and answer the following multiple choice question:

The following are 32 frames of the Video 1:

\texttt{<frame\_1> <frame\_2> ... <frame\_32>}

The following are 32 frames of the Video 2:

\texttt{<frame\_1> <frame\_2> ... <frame\_32>}

...

Q: \texttt{<question>}

\texttt{<options>}

Please select the correct answer from the options. Answer with the letter directly.

Your answer:
\end{tcolorbox}

\begin{tcolorbox}[title = {LLM as a Judge (Extract Selected Option)}]
  You judge which option a model selected for a multiple choice question.
  
  The question was:
  
  \texttt{<question>}
  
  Available options are: \texttt{<options>}
  
  The model's response was:
  
  \texttt{<model\_output>}
  
  Your task is to determine which option the model selected. Look for:
  
  \begin{itemize}\setlength{\itemsep}{0pt}
      \item Explicit mention of a letter (e.g., "A", "B", "C", "D")
      \item The model stating or implying a specific choice
      \item The response content matching one of the available options
  \end{itemize}
  
  If the model clearly selected one of the options, return \textbf{the corresponding letter}.
  
  If the model's response is empty, an error, unclear, or does not make a definitive choice, return \textbf{"X"}.
  \end{tcolorbox}

\begin{tcolorbox}[title = {Language Prior Filtering}]
You are answering multiple choice questions about video game footage.

You have NOT seen the video. Based only on the question and options provided, select the most likely answer.

You must respond with ONLY a single letter (A, B, C, or D).
\end{tcolorbox}

\section{Annotation Protocol}
\label{sec:annotation_protocol}

\subsection{Annotator Demographics and Expertise}
\label{sec:annotator_demographics}

Our annotation team consisted of 5 graduate students (ages 21--31; 3 male, 2 female), assigned to 4 labeler and 2 evaluator roles, with one participant serving in both capacities to ensure cross-stage consistency. Annotators were graduate student co-authors recruited internally for this study and were not financially compensated. All annotators were informed of the purpose of the data collection and consented to their annotations being used for research and potential public release.

\paragraph{General Gaming Experience.}
To confirm that annotators were qualified to interpret complex game states, we surveyed their overall gaming habits and game-specific familiarity before the annotation process.

\begin{itemize}
  \item \textbf{Video game play frequency in the past 100 days:} 60\% (3 participants) play regularly (3--5 times/week); 40\% (2 participants) play occasionally (1--2 times/week).
  \item \textbf{Years of video game experience:} 60\% (3 participants) have 8+ years; 20\% (1 participant) 3--8 years; 20\% (1 participant) 1--3 years.
\end{itemize}

\paragraph{Game-Specific Familiarity.}
Table~\ref{tab:annotator_familiarity} reports each annotator's self-reported familiarity with the benchmark titles.

\begin{table}[H]
  \centering
  \scriptsize
  \setlength{\tabcolsep}{3pt}
  \renewcommand{\arraystretch}{1.15}
  \begin{tabular}{lccccc}
  \toprule
  \textbf{Game} & \textbf{Expert} & \textbf{Regular} & \textbf{Casual} & \textbf{Low} & \textbf{None} \\
               & \textbf{(>300h)} & \textbf{(>30h)} & \textbf{(>5h)} & \textbf{(>0h)} & \\
  \midrule
  Counter-Strike  & 1 & 1 & 2 & 1 & 0 \\
  Minecraft       & 1 & 1 & 3 & 0 & 0 \\
  Apex Legends    & 0 & 2 & 0 & 3 & 0 \\
  Battlefield 6   & 0 & 2 & 0 & 3 & 0 \\
  Arc Raiders     & 0 & 1 & 0 & 4 & 0 \\
  Cyberpunk 2077  & 0 & 1 & 1 & 3 & 0 \\
  Elden Ring      & 0 & 1 & 1 & 2 & 1 \\
  Valheim         & 0 & 0 & 1 & 3 & 1 \\
  \bottomrule
  \end{tabular}
  \caption{Annotator familiarity with each game. Values indicate the number of annotators (out of 5) at each familiarity level. Hours denote minimum play time.}
  \label{tab:annotator_familiarity}
\end{table}

\subsection{Annotation Interface}
\label{sec:annotation_interface}

We developed a custom web-based annotation tool supporting both single-video and multi-video labeling. The annotation starts with \geminipro-generated timeline captions on the video about the six target entity types (SA, SS, OA, OS, WO, WE) and distractor candidates for lexical semantic distractors and scene distractors. Human annotators then verify each individual caption and distractor regarding type, content, and timeline. The annotation tool is publicly available:
\begin{itemize}
  \item \textbf{Source Code:} \url{https://github.com/wangyz1999/sync-video-label}
  \item \textbf{Demo Video:} \url{https://www.youtube.com/watch?v=PKedELJ4XT0}
  \item \textbf{Live Demo:} \url{https://sync-video-label.vercel.app/}
\end{itemize}

\begin{figure*}[t]
    \centering
    \includegraphics[width=0.99\linewidth]{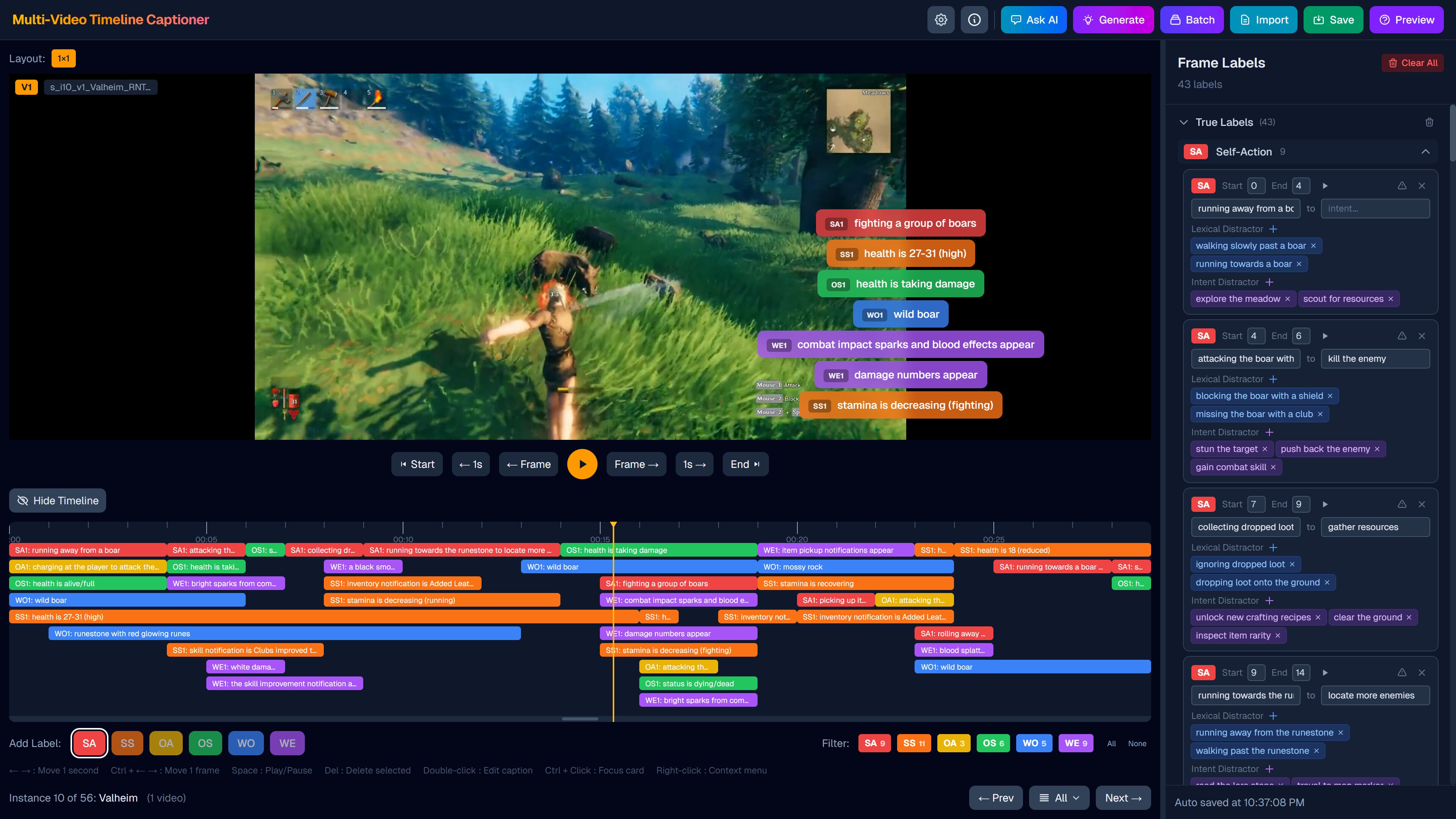}
    \caption{Screenshot of the annotation interface in the \textbf{single-video} setting, shown for the game \textit{Valheim}. Annotators label actions, states, events, and entities on a synchronized timeline.}
    \label{fig:interface_single}
\end{figure*}

\begin{figure*}[t]
    \centering
    \includegraphics[width=0.99\linewidth]{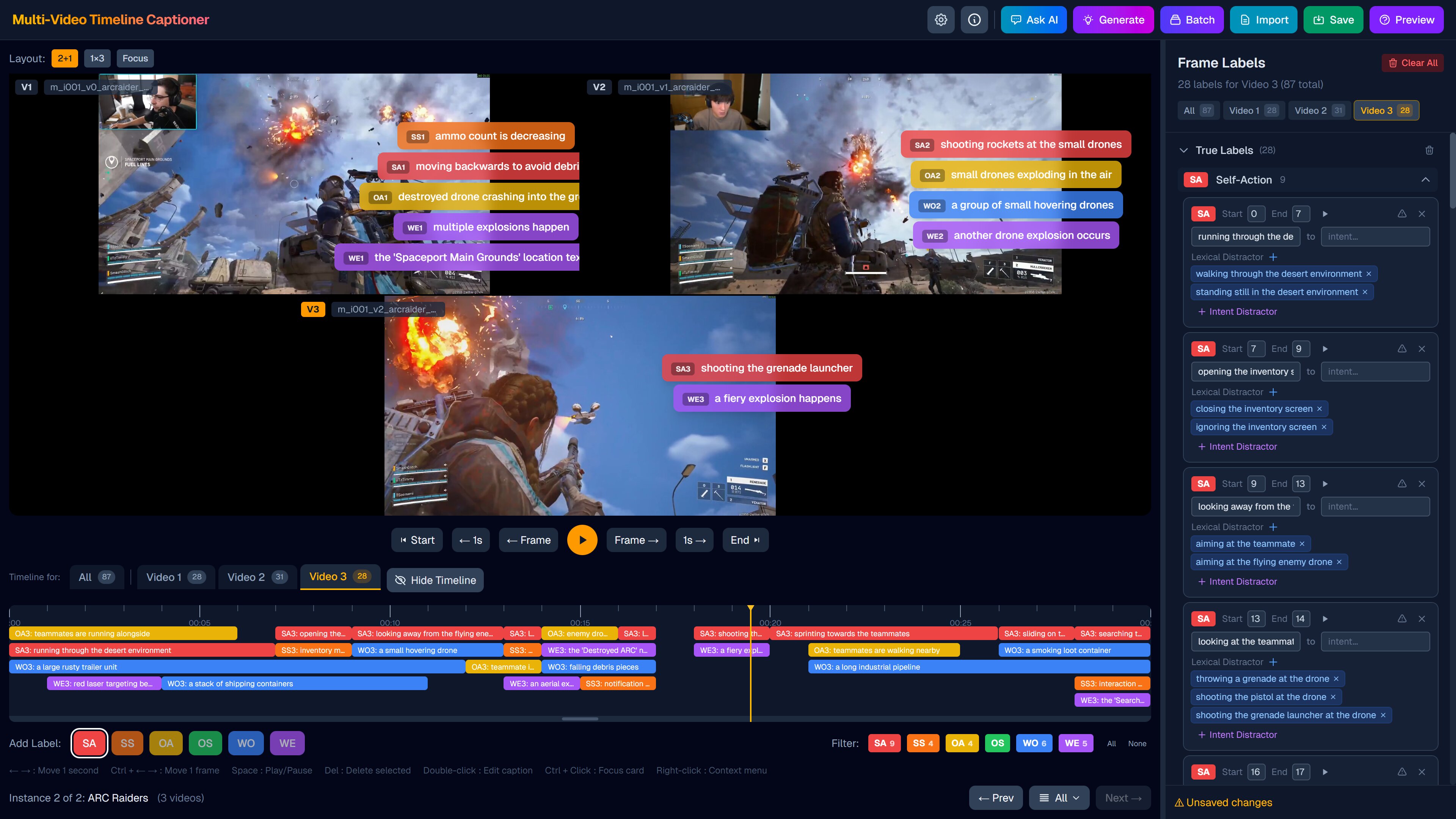}
    \caption{Screenshot of the annotation interface in the \textbf{multi-video} setting, shown for the game \textit{Arc Raiders}. Three synchronized video perspectives are displayed in parallel, with aligned timelines enabling annotators to capture cross-video events and temporal relations. The same explosion in the sky is captured by all three videos.}
    \label{fig:interface_multi}
\end{figure*}

Figure~\ref{fig:interface_single} shows an example of the interface in the single-video setting in the game \textit{Valheim}, where the player is combating a Boar. Figure~\ref{fig:interface_multi} shows an example of the interface in the multi-video setting in the game \textit{Arc Raiders}, where an explosion in the sky is synchronously captured by all three videos.

\subsection{Annotation Instructions}
\label{sec:annotation_instructions}

The annotation process follows a structured workflow designed to ensure temporal precision and semantic accuracy. Annotators begin by importing synchronized video instances (single or multi-POV) into the annotation interface. The process proceeds through three sequential phases: \textit{label generation}, \textit{verification}, and \textit{question preview}.

\textbf{Label Generation and Verification.} Annotators initiate automated caption generation using \geminipro, which produces candidate true labels, lexical distractors, and scene distractors for each video segment. For each generated true label, annotators verify three criteria: (1) the described event actually occurred, (2) the temporal boundaries $[t_{start}, t_{end}]$ accurately delimit the event, and (3) the assigned entity type (SA, SS, OA, OS, WO, WE) correctly categorizes the label. Temporal boundaries are adjusted when necessary to ensure precise alignment with observable events. For lexical distractors, annotators confirm that the described events do \textit{not} occur during the specified video segment. For scene distractors, verification ensures non-occurrence across the entire video duration.
Incorrectly generated labels are not immediately discarded; instead, we first consider whether they can be repurposed as distractors.

Overall, 31.1\% of \geminipro-predicted labels were deleted as incorrect or irreparable, while 42.7\% were edited to meet quality standards. Of these edited labels, 61.9\% required caption text corrections (e.g., fixing entity names, action descriptions, or semantic precision) and 42.2\% required temporal boundary adjustments to better align $[t_{start}, t_{end}]$ with the observable event. The remaining 26.2\% of predicted labels were accepted without modification. Additionally, 7.6\% of the final label set were added entirely by annotators to capture events missed by the model. In the second stage, a separate evaluator reviewed all labels and made further adjustments to approximately 12\% of labels, primarily to enforce cross-video consistency and resolve edge cases.

\textbf{Count Labeling.} Count questions require special handling: for action and event types (SA, OA, WE), annotators mark multiple temporally distinct segments sharing the same caption name. For object types (WO), annotators specify the quantity directly in the label metadata.

\textbf{Quality Control.} Annotators systematically review labels by filtering by entity type, verifying each category independently. Ambiguous labels, those where truth value cannot be reliably determined, are removed to maintain benchmark integrity. The interface supports efficient navigation through keyboard shortcuts and contextual menus, enabling rapid verification of temporal alignment and semantic correctness. Once all labels and distractors are verified, annotators proceed to the question preview phase, where generated questions are validated against the verified label set.

\section{Question Templates}
\label{sec:question_templates}

This section provides the complete list of question templates for all three levels used for question generation. Each template is identified by a code combining the entity type and question form. Placeholders: \plother{other} refers to other players (teammate, enemy, NPC); \plcaption{caption} refers to specific action/state/object/event descriptions; \plcaption{refCaption} refers to the anchor entity description; \plcaption{timestamp} refers to a formatted time range (e.g., \texttt{[00:01 to 00:12]}).

\begin{table*}[t]
\centering
\small
\setlength{\tabcolsep}{5pt}
\renewcommand{\arraystretch}{1.2}
\begin{tabular}{p{1.2cm}p{1cm}p{1.4cm}p{9.5cm}}
\toprule
\textbf{Form} & \textbf{Entity} & \textbf{Code} & \textbf{Template} \\
\midrule

\multirow{6}{*}{\textit{IDENT}}
& SA & \texttt{SA-IDENT} & Which of the following actions did the POV player perform during the video? \\
& SS & \texttt{SS-IDENT} & Which of the following best describes the POV player's state in the video? \\
& OA & \texttt{OA-IDENT} & Which of the following actions did \plother{other} perform during the video? \\
& OS & \texttt{OS-IDENT} & Which of the following best describes \plother{other}'s state in the video? \\
& WO & \texttt{WO-IDENT} & Which of the following objects appeared in the video? \\
& WE & \texttt{WE-IDENT} & Which of the following event occurred in the video? \\

\cmidrule{1-4}

\multirow{6}{*}{\textit{EXIST}}
& SA & \texttt{SA-EXIST} & Did the POV player perform the action: "\plcaption{caption}"? \\
& SS & \texttt{SS-EXIST} & Can you describe the POV player's state as: "\plcaption{caption}"? \\
& OA & \texttt{OA-EXIST} & Did the \plother{other} perform the action: "\plcaption{caption}"? \\
& OS & \texttt{OS-EXIST} & Can you describe the \plother{other}'s state as: "\plcaption{caption}"? \\
& WO & \texttt{WO-EXIST} & Did the object "\plcaption{caption}" appear in the video? \\
& WE & \texttt{WE-EXIST} & Did the event "\plcaption{caption}" occur in the video? \\

\cmidrule{1-4}

\multirow{6}{*}{\textit{ABSENT}}
& SA & \texttt{SA-ABSENT} & Which action did the POV player NOT perform? \\
& SS & \texttt{SS-ABSENT} & Which of the following states does not describe the POV player's state? \\
& OA & \texttt{OA-ABSENT} & Which action did the \plother{other} NOT perform? \\
& OS & \texttt{OS-ABSENT} & Which of the following does not describe the \plother{other}'s state? \\
& WO & \texttt{WO-ABSENT} & Which objects is NOT present in the scene? \\
& WE & \texttt{WE-ABSENT} & Which of the following events did NOT occur in the video? \\

\cmidrule{1-4}

\multirow{4}{*}{\textit{COUNT}}
& SA & \texttt{SA-COUNT} & How many times did the POV player perform the action: "\plcaption{caption}"? \\
& OA & \texttt{OA-COUNT} & How many times did the \plother{other} perform the action: "\plcaption{caption}"? \\
& WO & \texttt{WO-COUNT} & How many \plcaption{caption} are there in the scene? \\
& WE & \texttt{WE-COUNT} & How many times did the event "\plcaption{caption}" occur in the video? \\

\cmidrule{1-4}

\multirow{2}{*}{\textit{INTENT}}
& SA & \texttt{SA-INTENT} & Why did the POV player perform the action: "\plcaption{caption}"? \\
& OA & \texttt{OA-INTENT} & Why did the \plother{other} perform the action: "\plcaption{caption}"? \\

\bottomrule
\end{tabular}
\caption{Level 1 (perception) question templates. Entity types: SA (Self-Action), SS (Self-State), OA (Other-Action), OS (Other-State), WO (World-Object), WE (World-Event).}
\label{tab:question_templates_l1}
\end{table*}

\begin{table*}[t]
\centering
\small
\setlength{\tabcolsep}{4pt}
\renewcommand{\arraystretch}{1.2}
\begin{tabular}{p{1.3cm}p{1.5cm}p{2.2cm}p{8.5cm}}
\toprule
\textbf{Form} & \textbf{Ref\,→\,Ans} & \textbf{Code} & \textbf{Template} \\
\midrule

\multirow{6}{*}{\textit{IDENT}}
& SA\,→\,SS & \texttt{SA2SS-IDENT} & When the POV player was performing the action: ``\plcaption{refCaption}'', which of the following best describes their state? \\
& SA\,→\,OA & \texttt{SA2OA-IDENT} & When the POV player was performing the action: ``\plcaption{refCaption}'', which of the following actions did \plother{other} perform? \\
& SS\,→\,SA & \texttt{SS2SA-IDENT} & When the POV player's ``\plcaption{refCaption}'', which of the following actions did they perform? \\
& OA\,→\,SS & \texttt{OA2SS-IDENT} & When \plother{other} was performing the action: ``\plcaption{refCaption}'', which of the following best describes the POV player's state? \\
& WO\,→\,SA & \texttt{WO2SA-IDENT} & At the moment when the object ``\plcaption{refCaption}'' appeared, which of the following actions did the POV player perform? \\
& WE\,→\,OA & \texttt{WE2OA-IDENT} & At the moment when the event ``\plcaption{refCaption}'' occurred, which of the following actions did \plother{other} perform? \\

\cmidrule{1-4}

\multirow{6}{*}{\textit{EXIST}}
& SA\,→\,SS & \texttt{SA2SS-EXIST} & When the POV player was performing the action: ``\plcaption{refCaption}'', can you describe their state as: ``\plcaption{caption}''? \\
& SA\,→\,OA & \texttt{SA2OA-EXIST} & When the POV player was performing the action: ``\plcaption{refCaption}'', did \plother{other} perform the action: ``\plcaption{caption}''? \\
& SS\,→\,SA & \texttt{SS2SA-EXIST} & When the POV player's ``\plcaption{refCaption}'', did they perform the action: ``\plcaption{caption}''? \\
& OA\,→\,SS & \texttt{OA2SS-EXIST} & When \plother{other} was performing the action: ``\plcaption{refCaption}'', can you describe the POV player's state as: ``\plcaption{caption}''? \\
& WO\,→\,SA & \texttt{WO2SA-EXIST} & At the moment when the object ``\plcaption{refCaption}'' appeared, did the POV player perform the action: ``\plcaption{caption}''? \\
& WE\,→\,OA & \texttt{WE2OA-EXIST} & At the moment when the event ``\plcaption{refCaption}'' occurred, did \plother{other} perform the action: ``\plcaption{caption}''? \\

\cmidrule{1-4}

\multirow{6}{*}{\textit{ABSENT}}
& SA\,→\,SS & \texttt{SA2SS-ABSENT} & When the POV player was performing the action: ``\plcaption{refCaption}'', which of the following does NOT describe their state? \\
& SA\,→\,OA & \texttt{SA2OA-ABSENT} & When the POV player was performing the action: ``\plcaption{refCaption}'', which action did \plother{other} NOT perform? \\
& SS\,→\,SA & \texttt{SS2SA-ABSENT} & When the POV player's ``\plcaption{refCaption}'', which action did they NOT perform? \\
& OA\,→\,SS & \texttt{OA2SS-ABSENT} & When \plother{other} was performing the action: ``\plcaption{refCaption}'', which of the following does NOT describe the POV player's state? \\
& WO\,→\,SA & \texttt{WO2SA-ABSENT} & At the moment when the object ``\plcaption{refCaption}'' appeared, which action did the POV player NOT perform? \\
& WE\,→\,OA & \texttt{WE2OA-ABSENT} & At the moment when the event ``\plcaption{refCaption}'' occurred, which action did \plother{other} NOT perform? \\

\bottomrule
\end{tabular}
\caption{Level 2 entity-reference question templates (representative examples; full set covers all 30 Ref\,→\,Ans pairs across SA, SS, OA, OS, WO, WE). Code format: \texttt{\{Ref\}2\{Ans\}-\{Form\}}. Each EXIST question is further instantiated with distractor subtypes: \textit{True}, \textit{Lexical}, \textit{Scene}, \textit{Temporal}, and \textit{Role}.}
\label{tab:question_templates_l2_entity}
\end{table*}

\begin{table*}[t]
\centering
\small
\setlength{\tabcolsep}{4pt}
\renewcommand{\arraystretch}{1.2}
\begin{tabular}{p{1.3cm}p{1cm}p{2.0cm}p{9.2cm}}
\toprule
\textbf{Form} & \textbf{Ans} & \textbf{Code} & \textbf{Template} \\
\midrule

\multirow{6}{*}{\textit{IDENT}}
& SA & \texttt{TR2SA-IDENT} & During \plcaption{timestamp}, which of the following actions did the POV player perform? \\
& SS & \texttt{TR2SS-IDENT} & During \plcaption{timestamp}, which of the following best describes the POV player's state? \\
& OA & \texttt{TR2OA-IDENT} & During \plcaption{timestamp}, which of the following actions did \plother{other} perform? \\
& OS & \texttt{TR2OS-IDENT} & During \plcaption{timestamp}, which of the following best describes \plother{other}'s state? \\
& WO & \texttt{TR2WO-IDENT} & During \plcaption{timestamp}, which of the following objects appeared? \\
& WE & \texttt{TR2WE-IDENT} & During \plcaption{timestamp}, which of the following events occurred? \\

\cmidrule{1-4}

\multirow{6}{*}{\textit{EXIST}}
& SA & \texttt{TR2SA-EXIST} & During \plcaption{timestamp}, did the POV player perform the action: ``\plcaption{caption}''? \\
& SS & \texttt{TR2SS-EXIST} & During \plcaption{timestamp}, can you describe the POV player's state as: ``\plcaption{caption}''? \\
& OA & \texttt{TR2OA-EXIST} & During \plcaption{timestamp}, did \plother{other} perform the action: ``\plcaption{caption}''? \\
& OS & \texttt{TR2OS-EXIST} & During \plcaption{timestamp}, can you describe \plother{other}'s state as: ``\plcaption{caption}''? \\
& WO & \texttt{TR2WO-EXIST} & During \plcaption{timestamp}, did the object ``\plcaption{caption}'' appear? \\
& WE & \texttt{TR2WE-EXIST} & During \plcaption{timestamp}, did the event ``\plcaption{caption}'' occur? \\

\cmidrule{1-4}

\multirow{6}{*}{\textit{ABSENT}}
& SA & \texttt{TR2SA-ABSENT} & During \plcaption{timestamp}, which action did the POV player NOT perform? \\
& SS & \texttt{TR2SS-ABSENT} & During \plcaption{timestamp}, which of the following does NOT describe the POV player's state? \\
& OA & \texttt{TR2OA-ABSENT} & During \plcaption{timestamp}, which action did \plother{other} NOT perform? \\
& OS & \texttt{TR2OS-ABSENT} & During \plcaption{timestamp}, which of the following does NOT describe \plother{other}'s state? \\
& WO & \texttt{TR2WO-ABSENT} & During \plcaption{timestamp}, which object did NOT appear? \\
& WE & \texttt{TR2WE-ABSENT} & During \plcaption{timestamp}, which of the following events did NOT occur? \\

\bottomrule
\end{tabular}
\caption{Level 2 timestamp-reference (TR) question templates. The placeholder \plcaption{timestamp} is replaced by a formatted time range such as \texttt{[00:01 to 00:12]}. Code format: \texttt{TR2\{Ans\}-\{Form\}}.}
\label{tab:question_templates_l2_tr}
\end{table*}

\begin{table*}[t]
\centering
\small
\setlength{\tabcolsep}{4pt}
\renewcommand{\arraystretch}{1.2}
\begin{tabular}{p{1.5cm}p{1.5cm}p{2.8cm}p{7.7cm}}
\toprule
\textbf{Type} & \textbf{Ref\,→\,Ans} & \textbf{Code} & \textbf{Template} \\
\midrule

\multicolumn{4}{l}{\textit{Cross-Video Reference (V1-\{Ref\}2V2-\{Ans\}-\{Form\})}} \\
\cmidrule{1-4}

\multirow{4}{*}{\textit{IDENT}}
& SA\,→\,SA & \texttt{V1-SA2V2-SA-IDENT} & When POV1 player was performing the action: ``\plcaption{refCaption}'', which of the following actions did POV2 player perform at the same time? \\
& SA\,→\,SS & \texttt{V1-SA2V2-SS-IDENT} & When POV1 player was performing the action: ``\plcaption{refCaption}'', which of the following best describes POV2 player's state at the same time? \\
& OA\,→\,SA & \texttt{V1-OA2V2-SA-IDENT} & When \plother{refOther} was performing the action: ``\plcaption{refCaption}'' in POV1, which of the following actions did POV2 player perform at the same time? \\
& WE\,→\,WO & \texttt{V1-WE2V2-WO-IDENT} & At the moment when the event ``\plcaption{refCaption}'' occurred in POV1, which of the following objects appeared in POV2 at the same time? \\

\cmidrule{1-4}

\multirow{4}{*}{\textit{EXIST}}
& SA\,→\,SA & \texttt{V1-SA2V2-SA-EXIST} & When POV1 player was performing the action: ``\plcaption{refCaption}'', did POV2 player perform the action: ``\plcaption{caption}'' at the same time? \\
& SA\,→\,OA & \texttt{V1-SA2V2-OA-EXIST} & When POV1 player was performing the action: ``\plcaption{refCaption}'', did \plother{other} perform the action: ``\plcaption{caption}'' in POV2 at the same time? \\
& OS\,→\,WE & \texttt{V1-OS2V2-WE-EXIST} & When \plother{refOther}'s ``\plcaption{refCaption}'' in POV1, did the event ``\plcaption{caption}'' occur in POV2 at the same time? \\
& WO\,→\,SS & \texttt{V1-WO2V2-SS-EXIST} & When the object ``\plcaption{refCaption}'' appeared in POV1, was POV2 player's ``\plcaption{caption}'' at the same time? \\

\cmidrule{1-4}

\multicolumn{4}{l}{\textit{POV Identity (POV-ID): Which video corresponds to the player who did X?}} \\
\cmidrule{1-4}

\multirow{6}{*}{\textit{POV-ID}}
& SA & \texttt{SA-POV-ID} & Which video corresponds to the player who performed the action: ``\plcaption{caption}''? \\
& SS & \texttt{SS-POV-ID} & Which video corresponds to the player whose ``\plcaption{caption}''? \\
& OA & \texttt{OA-POV-ID} & Which video shows \plother{other} performing the action: ``\plcaption{caption}''? \\
& OS & \texttt{OS-POV-ID} & Which video shows \plother{other} whose ``\plcaption{caption}''? \\
& WO & \texttt{WO-POV-ID} & Which video shows the object ``\plcaption{caption}''? \\
& WE & \texttt{WE-POV-ID} & Which video shows the event ``\plcaption{caption}''? \\

\cmidrule{1-4}

\multicolumn{4}{l}{\textit{Temporal Ordering (ORDER): Which happened first?}} \\
\cmidrule{1-4}

\textit{ORDER} & SA & \texttt{SA-ORDER} & Which of the following actions happened first? \\

\bottomrule
\end{tabular}
\caption{Level 3 (cross-video) question templates. Cross-video reference templates cover all 36 Ref\,→\,Ans pairs (6 reference types $\times$ 6 answer types) in both IDENT and EXIST forms; representative examples are shown. \texttt{V1}/\texttt{V2} are replaced by actual video indices at generation time. \plother{refOther} refers to the other player in the reference video. POV-ID answer options are video numbers; ORDER options are formatted as ``The POV player in Video $X$ is [action]''.}
\label{tab:question_templates_l3}
\end{table*}

\end{document}